\documentclass[10pt,twocolumn,letterpaper]{article}
\usepackage[pagenumbers]{cvpr} 
\usepackage{graphicx}
\usepackage{amsmath}
\usepackage{amssymb}
\usepackage{booktabs}
\usepackage{graphicx}
\usepackage{wrapfig}
\usepackage{makecell}
\usepackage{xcolor}         
\usepackage{pifont}
\newcommand{\cmark}{\ding{51}}%
\newcommand{\xmark}{\ding{55}}%
\usepackage[font={small}]{caption}
\usepackage{mathtools}
\usepackage{mathtools, cuted}
\usepackage[misc]{ifsym}
\definecolor{antiquefuchsia}{rgb}{0.57, 0.36, 0.51}

\usepackage{amsmath}

\DeclareMathOperator*{\argmin}{arg\,min}
\usepackage{multirow}
\usepackage{balance}
\usepackage[ruled,vlined]{algorithm2e}
\definecolor{citecolor}{RGB}{65,105,225}
\usepackage[pagebackref=true,breaklinks=true,letterpaper=true,colorlinks,citecolor=citecolor,bookmarks=false]{hyperref}

\usepackage[capitalize]{cleveref}
\crefname{section}{Sec.}{Secs.}
\crefname{section}{Section}{Sections}
\Crefname{table}{Table}{Tables}
\crefname{table}{Tab.}{Tabs.}

\begin{document}
\title{Generalizable Neural Performer: Learning Robust Radiance Fields for\\Human Novel View Synthesis}

\author{Wei Cheng$^{1}$
    \quad
    Su Xu$^{1}$
    \quad
    Jingtan Piao$^{1,2}$ 
    \quad
    Chen Qian$^{1}$  \\
     \quad
    Wayne Wu$^{1,3}$
    \quad
    Kwan-Yee Lin\textsuperscript{1,2 \Letter}
    \quad
    Hongsheng Li\textsuperscript{2 \Letter}\\
    $^{1}$ SenseTime Research
    \quad
    $^{2}$ CUHK-SenseTime Joint Laboratory, CUHK
    \\
    $^{3}$ Shanghai AI Laboratory
    \\
    {\tt\small \{chengwei,xusu,piaojingtan,qianchen\}@sensetime.com}
    \\
    {\tt\small \{wuwenyan0503,linjunyi9335\}@gmail.com}
    \quad
    {\tt\small hsli@ee.cuhk.edu.hk}
}

\twocolumn[{
  \renewcommand\twocolumn[1][]{#1}
  \maketitle
  \begin{center}
  \vspace{-0.75cm}
  \includegraphics[width=0.9\textwidth]{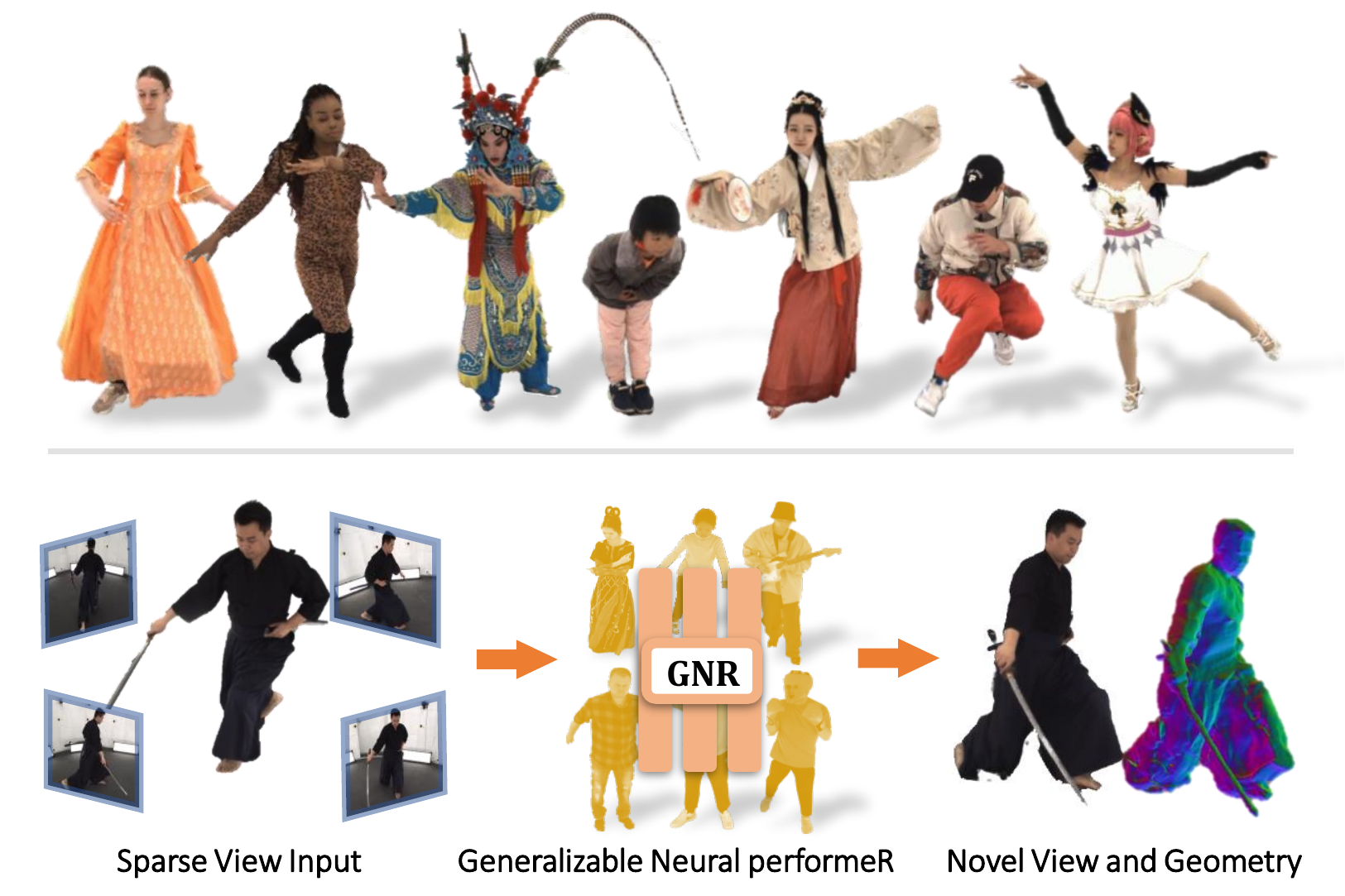}
  \captionof{figure}{\small{\textbf{Overview.} First row presents some snapshots in our dataset, GeneBody$-1.0$. It is a multi-view dataset which captures human motion in a variety of pose actions, body shape, clothing and accessories. Second row shows the simplification of our framework.}.}
  \label{fig:intro-large}
  \end{center}
}]


\begin{abstract}

This work targets at using a general deep learning framework to synthesize free-viewpoint images of arbitrary human performers, only requiring a sparse number of camera views as inputs and skirting per-case fine-tuning. The large variation of geometry and appearance, caused by articulated body poses, shapes and clothing types, are the key bottlenecks of this task. To overcome these challenges, we present a simple yet powerful framework, named Generalizable Neural Performer (GNR), that learns a generalizable and robust neural body representation over various geometry and appearance. Specifically, we compress the light fields for novel view human rendering as conditional implicit neural radiance fields with several designs from both geometry and appearance aspects. We first introduce an Implicit Geometric Body Embedding strategy to enhance the robustness based on both parametric 3D human body model prior and multi-view source images' hints. On the top of this, we further propose a Screen-Space Occlusion-Aware Appearance Blending technique to preserve the high-quality appearance, through interpolating source view appearance to the radiance fields with a relax but approximate geometric guidance.
To evaluate our method, we present our on-going effort of constructing a dataset with remarkable complexity and diversity. The first dataset version, GeneBody-1.0, includes over $2.95M$ frames of $100$ subjects under multi-view cameras capturing, performing a large variety of pose actions, along with diverse body shapes, clothing, accessories and hairdos. Experiments on GeneBody-1.0 and ZJU-Mocap show better robustness of our methods than recent state-of-the-art generalizable methods among all cross-dataset, unseen subjects and unseen poses settings. We also demonstrate the competitiveness of our model compared with cutting-edge case-specific ones. Dataset, code and model will be made publicly available\footnote{Project page: \url{https://generalizable-neural-performer.github.io/}\\
Code and model: \url{https://github.com/generalizable-neural-performer/gnr}}.
\end{abstract}


\section{Introduction}\label{intro}
%
\begin{figure*}[!t]
\centering
\includegraphics[width=0.95\textwidth]{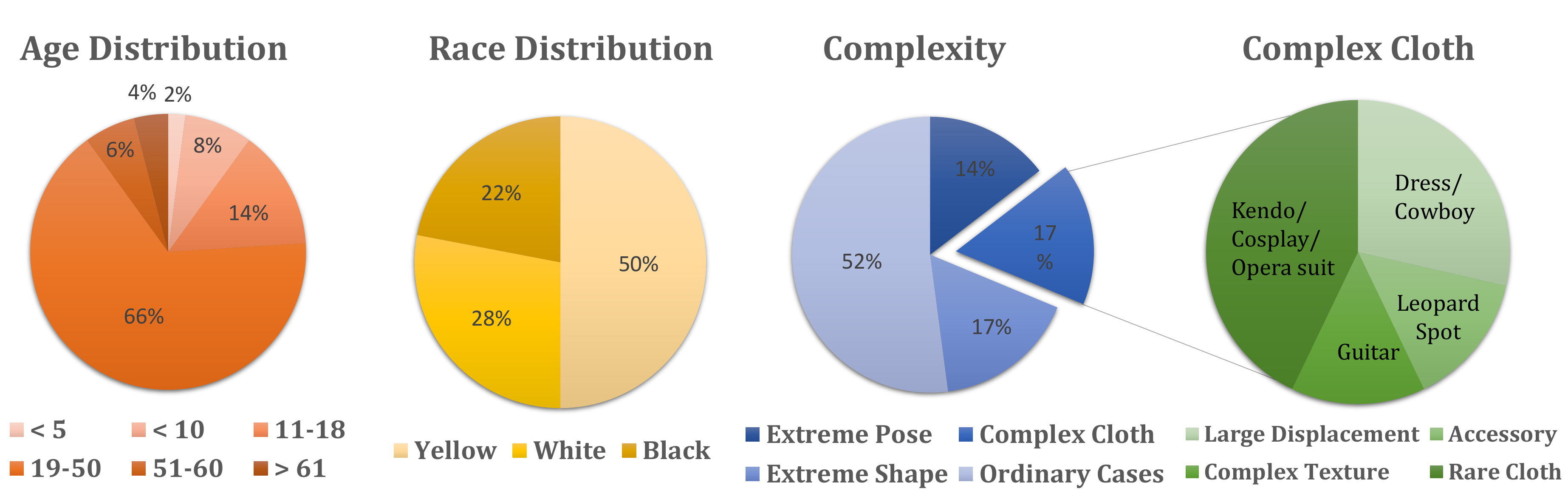}
\vspace{-2ex}
\caption{\small \textbf{Main Statistics of GeneBody-1.0.} It contains wide range over ages and races. The complexity aspect comes from analyzing the dominant challenging aspect of all cases, and exhibits detailed samples attributed to \emph{Complex Cloth}.}\label{fig:distribution}
\end{figure*}

Synthesizing free-viewpoint images of human performers has a wide range of applications in film production, 3D immersive communication and AR/VR gaming. However, the most impressive industry-level rendering results currently require both specialized studio environments~\cite{2007The} and cumbersome artistic design~\cite{web:unrealengine} on each human subject. What if such stirring effect could be automatically realized with just using images captured under a casual setting (e.g., very limited number of cameras)? Such capability would dramatically accelerates filmdom and improves the accessibility of 3D immersive experience in daily life.

In this work, we focus on improving the \textit{generalization} and \textit{robustness} in free-viewpoint synthesis for arbitrary human performers from a sparse set of multi-view images. To achieve this, two key challenges need to be solved. First, for the generalization of unseen subjects and unseen actions during inference, the model is asked to represent arbitrary shape and appearance variation caused by different human posing and clothing.
Second, rendering high-quality results requires detail preserving of appearance as well as multi-view consistency.

Many previous pioneer works fall short of these goals. There has been attempts either rely on deploying depth sensors for high-quality geometry reconstruction before rendering~\cite{guo2017real,dou2016fusion4d}, or building dense camera arrays to capture the changing appearances from different viewing angles~\cite{gortler1996lumigraph}. The requirement of professional equipment limits the application of such technologies in personal and daily usage scenarios.
Recent works adopt neural networks to learn 3D geometry and appearance from data~\cite{saito2019pifu,zheng2021pamir,kwon2021neural,peng2020neural}. Among these methods, the neural implicit representations have show efficient improvements over image fidelity and reconstruction accuracy. 
Despite the notable technical revolution they bring in, these methods still suffer from case-specific optimization~\cite{peng2020neural}, scanty pose generalization~\cite{saito2019pifu,saito2020pifuhd}, or unrealistic rendering~\cite{saito2019pifu,zheng2021pamir,kwon2021neural}. Some current cutting-edge approaches rely on temporal coherence of the same subject~\cite{kwon2021neural,peng2020neural} which requires geometry fitting or motion tracking across canonical models. Although lavish geometry and appearance cues lies in 3D body motion across time, there are more instantaneous snapshots sources rather videos in daily life. 

To address aforementioned challenges, we present \textit{\textbf{G}eneralizable \textbf{N}eural Performe\textbf{R}}, or GNR for short, a novel framework that learns a generalizable and robust neural body representation on various human pose, shape and appearance. We achieve this with several effective designs on tailoring body structural prior and source-view appearance conditions into neural radiance fields learning.

Specifically, to be able to represent arbitrary human performers in one single model without per-case finetuning, an \textit{Implicit Geometric Body Embedding} strategy is firstly introduced. It extracts the geometric information and body semantics from both parametric 3D human body model and multi-view source image feature fusion, which turns the neural radiance field into a conditional implicit field. This strategy has the following properties: $1)$ The existing parametric models mostly represent ``minimally clothed'' bodies (e.g., SMPL~\cite{loper2015smpl}), which could provide basic geometric prior in pose and constrain the minimal shape. $2)$The multi-view input image features encode the firsthand geometric hints of hairdo, cloth and accessories that form the unabridged shape. Thus, the predicted human geometry could be conditionally anchored\footnote{We use the word ~`anchor` to indicate constraining the inferred human shape to be close to the proper body surface.} by the parametric body prior and further bordered by multi-view observation. In short, such a embedding could facilitate network learning a reasonable body pose and shape parameterization via a conditional implicit neural representation for articulated clothed human and thus skirting per-scene optimization. A proper geometry prediction could also in turn helps the generalization of appearance. 

\begin{table*}[!t]
\small
\begin{center}
\resizebox{0.95\textwidth}{!}{
\begin{tabular}{c|ccccccc}
\Xhline{1.0pt}
\textbf{Methods}  & \textbf{Generalizable} & \textbf{Render} & \textbf{Latent} & \textbf{Prior} &\textbf{Occlusion} & \textbf{Supervision}\\
\Xhline{1.0pt}
NeRF\cite{mildenhall2020nerf} & \xmark & \cmark & - & - & - &2D\\
pixelNerf\cite{yu2020pixelnerf}& \cmark & \cmark & 2D & - &- & 2D\\
IBRNet\cite{wang2021ibrnet}& \cmark & \cmark & 2D & - & - &2D\\
PIFu\cite{saito2019pifu} & \cmark & \xmark & 2D & -  & - &3D\\
PaMIR\cite{zheng2021pamir} & \cmark & \xmark & 2D+3D & SMPL (Depth) &- & 3D\\
NB\cite{peng2020neural} & \xmark & \cmark & 2D+3D & SMPL (Vertices) &- & 2D\\
\Xhline{1.0pt}
GNR  & \cmark & \cmark & 2D+3D & SMPL (SDF\&Depth) & \cmark  &2D($\text{3D}^{*}$)\\
\Xhline{1.0pt}
\end{tabular}
}
\end{center}
\caption{\small \textbf{Summary of key related methodology.} \textbf{Generalizable}: Ability to model multiple objects/scenes in same model. \textbf{Render}: Whether has a learnable render. \textbf{Latent}: Latent code types used to condition the network. \textbf{Prior}: Type of prior used to bound the structure of objects/scenes, eg., SMPL~\cite{loper2015smpl} as human body piror. \textbf{Occlusion}: Whether consider occlusion during rendering. \textbf{Supervision}: Type of losses. $*$ indicates the type of the supervision is optional. }\label{tab:compare}
\end{table*}

To robustly producing faithful appearance, we further propose an {\textit{Screen-Space Occlusion-Aware Appearance Blending (SSOA-AB)}} technique. It combines lumigraph blending techniques~\cite{gortler1996lumigraph,kellnhofer2021neural} with screen space occlusion estimation. The key aspect of this design is to disentangle blending coefficient from source views into a \textit{visibility map}, guided by parametric model prior and screen-space self-occlusion information, and a {\textit {view attention coefficient}} between source and query views learnt from large-scale training data prior. 
As the texture from source views provides authentic details, the pixel from these views could be treated as the firsthand base reference color to rectify the unreliable color prediction of the neural radiance field. Whereas not all points could be observed through input views due to sparse view observation as well as occlusions of body parts, directly applying classic lumigraph blending will lead to ghosting texture in fully unobserved part. 
We thus introduce SSOA-AB, which blends from both source views and learned radiance itself, by treating the radiance as the observation from a virtual camera, and blends these according to learnt visibility and score.
Such design will help with a more proper blending, which result in reasonable appearance detail preserving and multi-view consistency.

Recent datasets~\cite{Joo_2017_TPAMI,peng2020neural,ionescu2013human3,yu2020humbi} are inadequate to train or evaluate an effective model for synthesising human in real-world scenarios, which owns complex and diverse geometry and appearance. Thus, we contribute a dataset called GeneBody-1.0. It consists of over $4M$ frames of $100$ subjects under multi-view cameras capturing, performing a variety of pose actions, in different types of body shape, clothing, accessories and hairdos, ranging the geometry and appearance varies from everyday life to professional occasions.
Some examples are shown in Fig.~\ref{fig:intro-large} and main dimensions of dataset construction are shown in Fig.~\ref{fig:distribution}. We use both synthetic dataset RenderPeople~\cite{web:renderpeople} and our real-world GeneBody dataset to train our model, yielding significant performance improvement in rendering and reconstruction over recent state-of-the-art generalizable volume rendering methods. Surprisingly, our generalized model achieves better rendering qualities than cutting-edge case specific methods in some cases with challenging clothes or poses, which further demonstrates robustness of proposed designs.

To summarize, our work contributes as follows
\vspace{-0.2cm}
\begin{itemize}
    \item We propose a novel approach, named Generalizable Neural Performer, that achieves free-viewpoint synthesis of \emph{arbitrary} human performers with learning a generalizable and robust implicit neural body representation. 

    \item We present an Implicit Geometric Body Embedding strategy and a Screen-Space Occlusion-Aware Blending technique to facilitate the learning of the representation. We show that a proper design, like these two components, for tailoring prior knowledge of parametric model and source images, can effectively enhance robustness to both human geometry and appearance. 
    
    \item We contribute a multi-view dataset, GeneBody-1.0, with human samples under various appearances on consideration of ethnicity, age, clothing style and conducting with performing diverse types of actions. We hope this dataset facilitate future research in generalizable human rendering towards real-world scenarios. We also benchmark results of several state-of-the-arts, with hope to offer some insights into current status and future trends in this field.
    
\end{itemize}

\section{Related Work}
\subsection{Novel View Rendering} 

\noindent \textbf{Image-based Rendering.}
Image-based rendering (IBR) is the fundamental problem of rendering novel views of scenes from sampled views. Plenoptic sampling theorem~\cite{chai2000plenoptic} determines minimum sampling rate for anti-aliased rendering bounded by minimum and maximum scene depths. Methods based on this theorem require dense view sampling to achieve high-quality rendering; many systems devise large-scale camera arrays~\cite{wilburn2005high}. Other methods explored to use geometric proxies, like global meshes~\cite{wood2000surface,hedman2018instant,buehler2001unstructured} or local layered depth~\cite{zhou2018stereo,mildenhall2019local} to alleviate sampling requirement. While the performance of these methods highly relies on the accuracy of geometry estimation and typically fails on low-textured regions.

\begin{figure*}[ht]
\centering
\includegraphics[width=.98\linewidth]{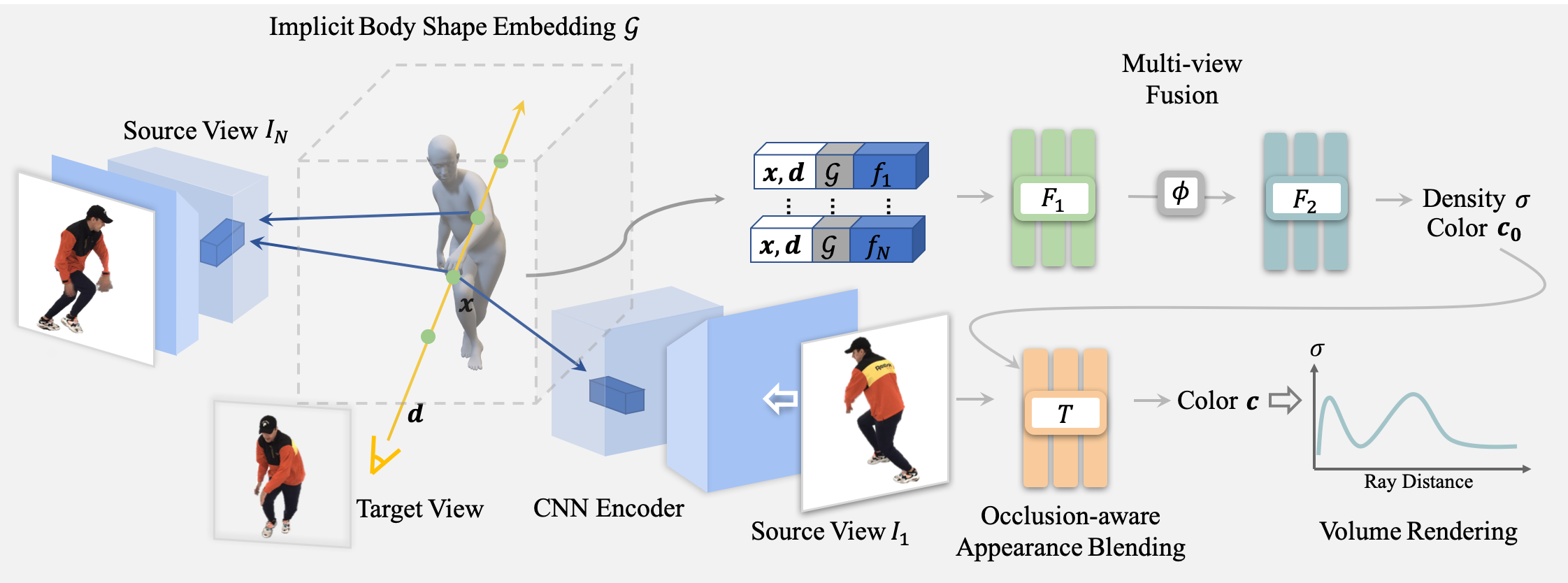}
\vspace{-1.5ex}
\caption{\small{\textbf{Framework.} Given a sparse set of multi-view images of a human performer along with his/her corresponding parametric 3d body model(e.g. SMPLx~\cite{pavlakos2019expressive}), we synthesize images by ray casting and sampling from target views. For each sampled point $\mathbf{x}$, its geometric information is anchored by both parametric model and fused multi-view features from source views, which then constitutes the features of Implicit Geometric Body Embedding to condition the neural radiance field. In parallel, GNR blends appearances from source views and predicted color $\mathbf{c}_0$ under the guidance of the occlusion map through the screen-space occlusion-aware appearance blending module. Novel view images are finally rendered via ray integration.}} 
\vspace{-3ex}
\end{figure*}

\noindent \textbf{Neural Scene Representation.}
Recent works use neural network to represent a scene~\cite{mildenhall2020nerf,saito2019pifu}, although these methods can generate rather satisfying novel views using scene-specific models, the quality of rendered images tends to degrade when model is trained to fit various objects/scenes. Recent works pixelNerf~\cite{yu2020pixelnerf} and IBRNet~\cite{wang2021ibrnet} improves generalization of 3D implicit functions by conditioning query coordinates with image features. These methods are typically unable to generate plausible free-viewpoint video with significant camera movement.

\subsection{Human Digitization} 

\noindent \textbf{Deep Representation with Parametric Models.}
The emergence of statistical models like SCAPE~\cite{anguelov2005scape}, and SMPL~\cite{loper2015smpl} create another cue of modeling humans in a general way, which is also treated as initial guesses in recent works.
~\cite{raj2020anr} estimates parametric model from video and renders the full model by per-frame silhouette. ~\cite{zheng2021pamir,huang2020arch} extract 3D features from voxelized SMPL to condition implicit function and achieves pose robustness compared to \cite{saito2019pifu}. While the requirement for a precise body scan hinders its robustness due to the biased data. 

\noindent \textbf{Modelling and Rendering Human from Images.}
To model/render human, previous study focus on reconstruction from shape~\cite{waschbusch2005scalable}, stereo~\cite{liu2009point} and shading~\cite{vlasic2009dynamic} using multi-view systems.
While these works are subject-specific and require sophisticated capturing volume setup and system calibration. Recently, ~\cite{saito2019pifu,saito2020pifuhd,zins2021learning,kwon2021neural} have made progress on reconstructing dense human models from a sparse set of images or even a single image in the wild by adopting data-driven techniques. 
Recently, some concurrent work in preprint~\cite{zhao2021humannerf,chen2021geometry,hnerf,kwon2021neural} generalize human in similar spirit.   
In contrast to these methods utilizing temporal information to complement the insufficiency of singe-frame sparse-view sources, our framework tries to generalize human without extra effort on temporal coherent. 

A summarized comparison of proposed method with most relevant related work is shown in Tab.~\ref{tab:compare}.

\section{Method}\label{system}

\noindent\textbf{Overview.}
Recall that given a sparse set of calibrated multi-view images  $\mathbf{I}=\{I_i\}_{i=1}^N$ of a person, our goal is to synthesize high-quality images of that person in arbitrary novel views. To this end, our model is asked to tackle challenges of $(1)$ representing arbitrary human performers in one single model without retraining on target case; $(2)$ producing faithful appearance with detail preserving and multi-view consistency. We alleviate these two dilemmas with introducing an Implicit Geometric Body Embedding strategy (for the former) and a Screen-Space Occlusion-aware Appearance Blending (for the latter) into the learning of a conditional neural radiance field $\mathcal{F}${\footnote{The basic principle of neural radiance field, or short NeRF \cite{mildenhall2020nerf}, is to learn a continuous 5D function by mapping a 3D location $\mathbf{x}=(x,y,z)$ within a bounding volume surrounding the scene and a 2D view direction $\mathbf{d}$ to an emitted color $\mathbf{c}=(r,g,b)$ and a volume density $\sigma$. The pixel is finally rendered via ray integration.}}

Specifically, we assume a parametric human body 
$M$,which fitted to input source images, is given as prior. A deft implicit body shape encoding $\mathcal{G}$ is built upon $M$ to anchor performer's pose and constrain the roughly occupied range in 3D space. In parallel, we extract source images’ features from the encoder ${E}$ and fuse them via multi-view feature fusion to form the unabridged shape hints complementing with $\mathcal{G}$. Since multi-view input image features encode the firsthand geometric cues of off-minimal body part like hairdo, cloth and accessories.
Then, for a query point $\mathbf{x}$ in 3d space, we look up the corresponding point-aligned features from both $G$ and $E$ to constitute the features of Implicit Geometric Body Embedding (Sec.~\ref{igbe}) to condition the implicit field for the radiance prediction. Alongside, we fed corresponding pixel color from source views and predicted color from the radiance field into the Screen-Space Occlusion-aware Appearance Blending module  (Sec.~\ref{app-blending}) to obtain the rectified color value, which is the final output color of GNR. Details are provided in ensuing subsections. 

\subsection{Implicit Geometric Body Embedding}\label{igbe}
\noindent\textbf{Implicit Body Shape Encoding.}\label{smpl-embedding}
It is intuitive to use a parametric body model as hands-down geometric prior to anchor human shape to the pose during the learning of implicit field. The SMPLx model $M(\theta,\beta)$ is a mesh consisting of $n_s = 10475$ vertices with fixed topology. $\theta\in\mathbb{R}^{72}$ and $\beta\in\mathbb{R}^{10}$ denote 
human pose in joints rotation and hyper-parameters of body shape from a statistic model. However, injecting a dense mesh or hyper-parameters of SMPLx is impractical and meaningless, due to its high dimensionality and requirement on shape basis.

We thus propose an alternative to use the signed distance function (SDF) of the mesh to parametrize the Euclidean volume.
Specifically, for any 3D location $\mathbf{x}$, we first compute its nearest point on the mesh surface $\mathbf{v} = \sum_{j=1}^3 c_j * \mathbf{v}_{i,j}$, 
where $\mathbf{v}_{i,j}$ is the $j$-th vertex on the nearest triangular surface $i$, and $c_j$ is the projected coefficients with constraints $\sum_{j=1}^3 c_j = 1,\ \ c_j \ge 0$. Next, the SDF of the query point $\mathbf{x}$ can be calculated as $S(\mathbf{x}, M) = {\rm sign}(\mathbf{x}, M) ||\mathbf{x} - \mathbf{v}||_2$, where ${\rm sign}(\mathbf{x}, M)$ equal to $1$ when $\mathbf{x}$ is inside $M$ and $-1$ when it is outside $M$.

Note that the $\rm sign$ function requires $M$ to be a water-tight mesh. This embedding benefits in two ways: 1) The SDF embedding provides information on how close the query point inwards or outwards the minimal human body surface. 2) Converting implicit SDF from mesh, the compact and efficient properties of 3D geometry are preserved, as demonstrated in~\cite{zheng2020deep,park2019deepsdf,mescheder2019occupancy}. Besides, we also incorporate
SDF's derivative with respect to $\mathbf{x}$, where we denote it as $S'_{\mathbf{x}}$. The derivative explains in which direction the point is approaching / leaving the mesh surface, or in other word, the normal direction of the surface.

Still and all, the SDF value and its derivative cannot reflect body part semantics, which would lead to local geometry,~i.e.,\ body part ambiguity, when the body pose changes. Therefore, to further eliminate such ambiguity, the body semantic knowledge is expected to be incorporated. That is, where on the body does the location of query point $\mathbf{x}$ approximately lie. We choose to map the closest point $\mathbf{v}$ on the SMPLx model $M(\theta, \beta)$ to its corresponding coordinates 
$\mathbf{\bar{v}}$ in a canonical SMPLx mesh $\bar{M}$ with neutral shape $\bar{\beta}$ and static pose $\bar{\theta}$, and $\bar{M} = M(\bar{\theta}, \bar{\beta})$.
Locations of the corresponding points in the canonical space serve as an effective representation of the semantic context regarding the template human body.
The complete body shape encoding is written as $ \mathcal{G}(M, x; \bar{M}) = [S; S'_{\mathbf{x}}; \mathbf{\bar{v}}]$,
where $[;]$ denotes concatenation. The proposed implicit body shape encoding helps to anchor the implicit radiance field to the proper human body locations in several ways:$(1)$\emph{Geometry generalization.} Introducing the dense parametric model information to the implicit field can be interpreted as providing an initial guess of the human pose and shape. The network is thereby trained with a easier task of learning the \textit{residual} between clothed human and its corresponding minimally clothed body shape. $(2)$ \emph{Low cost.} To provide meaningful constraints to the implicit field,
    conventional ways are either expensive in computation and memory as they directly incorporate voxelized body volume/ local 3D patch features~\cite{zheng2021pamir,peng2020neural,zins2021learning} , or lack geometry prior due to over-simplification of a relative direction to articulate skeleton representation~\cite{su2021nerf}. In contrast, our implicit body shape encoding combines local geometric prior and body semantics from body template, achieving excellent computation and memory efficiency. Such simple representation effectively conditions the implicit field. 

\begin{figure*}[t]
\centering
\includegraphics[width=.95\linewidth]{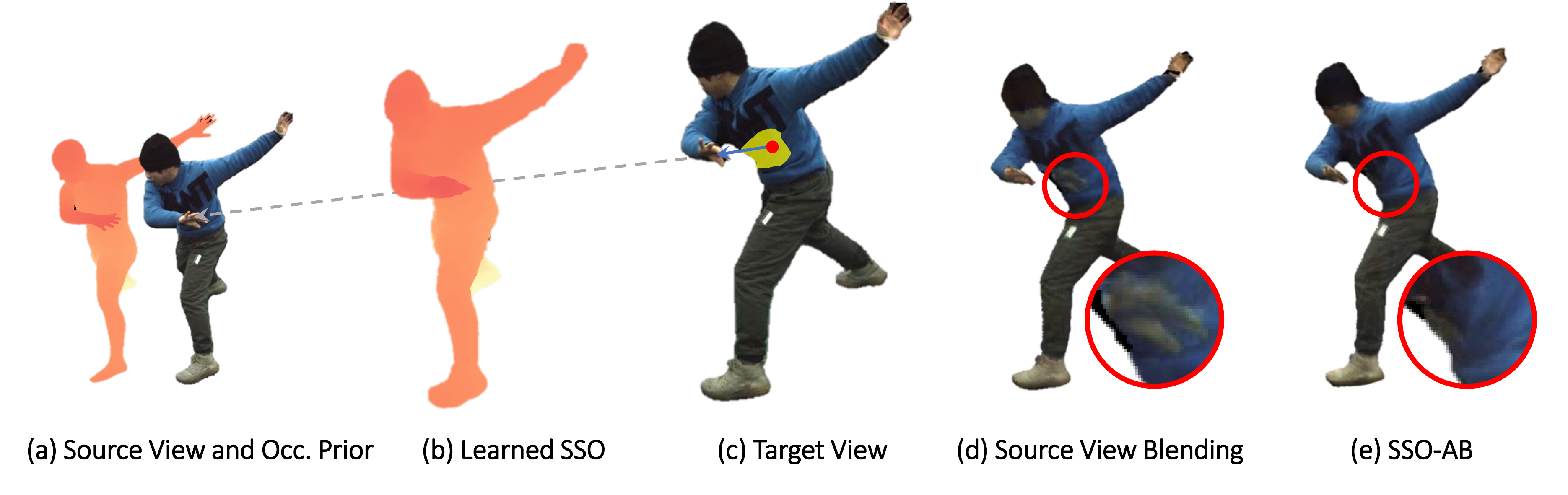}
\vspace{-2ex}
\caption{\small \textbf{Toy example of SSO-AB.} (a) Given a source view and its occlusion prior from parametric model, (b) a per-view screen-space occlusion is estimated during camera blending. (c) For novel view regions fully occluded from all source views (marked in yellow), (d) blending source views only will lead in ghosting artifact. \textbf{(e)} SSO-AB treat learned radiance as well as source view textures, enables ghosting-free textures.}\label{fig:att}
\vspace{-3ex}
\end{figure*}

\noindent\textbf{Multi-view Feature Fusions and Radiance Prediction.}\label{multiview}
As most parametric models represent ~`minimally clothed` bodies, simply condition the implicit filed with only implicit body shape encoding is insufficient to infer off-minimal body shape part,~\textit{e.g.,} hairdo, cloth and accessories. Thus, we need a complement to offer these cues. As the multi-view input image features intrinsically encode firsthand geometric information of these aspects, the implicit filed is also conditioned by the pixel-aligned features from multi-view inputs aside from implicit body shape encoding. 
Specifically, we first extract the feature using a convolution network, denoted as $E$ {\footnote{Note that we adopt the same network architecture with \cite{saito2019pifu} and initialize its parameters from the pretrained model.}}. 
The query point $\mathbf{x}$ is projected to 2D location $\mathbf{p}$ on each source image planes given the camera pose and intrinsic parameter, and the image feature vector is obtain via $f(\mathbf{x}) = E(\mathbf{p})$ with bilinearly interpolation.

We separate density and radiance reasoning network $\mathcal{F}_\sigma$, $\mathcal{F}_\mathbf{c_0}$ into two stages:
\begin{equation}
    \mathcal{F}_\sigma, \mathcal{F}_\mathbf{c_0} = F_2 \circ \phi [ F_1(\mathbf{x}, \mathcal{G}, f) ],
\label{eq:alpha-fusion}
\end{equation}
where $F_1$ is an MLP that takes $\mathbf{x}$, body model embedding $\mathcal{G}$ and image feature vector $f$ as input and outputs an intermediate feature embedding for each of source views. Like~\cite{saito2019pifu,yu2020pixelnerf}, these intermediate features are fused via average pooling $\phi$, and aggregated feature is then passed into the volume density and radiance reasoning MLP $F_2$ to predict occupancy probability and radiance of $\mathbf{x}$.

\subsection{Screen-Space Occlusion-aware Appearance Blending}\label{app-blending}
Rendering high-quality images of arbitrary human performer from a general implicit function without retraining on that case is difficult, especially at regions with high-frequency textures, {\textit{e.g.,}} hair, accessory and cloth textures. Rendering at these regions may be blurry or even glitchy. Recent works learn the camera blending weights of implicit scenes~\cite{wang2021ibrnet} or geometry proxies~\cite{gortler1996lumigraph} from source view images. However, these methods require dense view references based on the dense assumption of unstructured lumigraph rendering~\cite{buehler2001unstructured}, and fail in sparse setting with further 
triggering ghosting effect especially in source view occluded part. 
To solve this, our idea is to disentangle the camera blending field $\mathbb{B}$ into an occlusion-aware visibility $o$ and a view attention based blending coefficient $\gamma$.

\noindent\textbf{Screen-Space Occlusion.}
Ideally, we can infer the texture of any point once a precise occlusion of clothed human and dense source views are provided.
However, computing precise occlusion requires either casting rays from a certain point and checks for intersection with reconstructed mesh
or estimating depth from source view ray integration. 
To train such network is difficult and computational ineffective.
While luckily, the hands-down parametric body model could provide initial geometric information of self-occlusion and source-view image observations offer visual clue of depth value continuity in out-of-body regions.
Thus, we can instead learn an approximate occlusion in the observed screen-space with enforcing the network to diffuse the initial minimal occlusion to clothed ones.

Based on this, we propose a network $\mathcal{F}_o$ to learn the occlusion (Learned SSO in Fig~\ref{fig:att}(b)) from given  minimal-clothed occlusion map (Occ.Prior in Fig~\ref{fig:att}(a)) and image features, the occlusion visibility can be written as 
\begin{equation}
    o(\mathbf{x}, \mathbf{d}) = \mathcal{F}_o(\mathbf{x}, \mathbf{d}; e_{\mathbf{x},\mathbf{d}}, \mathcal{F}_1)
\end{equation}
where we model $e = \psi(z_{ref}-z)$ as the visibility of $\mathbf{x}$ from view direction ${\mathbf{d}}$ in context of SMPLx model, and $z_{ref}$ is the view depth of SMPLx model which can be obtained via a rasterization process, and $z$ is depth of $\mathbf{x}$, $\psi$ is a sigmoid.

\noindent\textbf{View Attention based Camera Blending.}
We also propose a view attention block to learn the appropriate view attention coefficient from angluar and visual similarity under large-scale data prior. Specifically, 
we deploy an attention module originated from transformer block~\cite{vaswani2017attention}, to model the camera blending field $T$ according to the self- and cross-similarities of learned features. 
$Q$ analogises to the query feature in the context of transformer, and $K$ analogises to the key. In our scenario, they are computed from query point $\mathbf{x}$, view directions $\mathbf{d}_i$ and view feature embedding $F_{1,i}$ through two separate MLPs. Our final color prediction process can be written as the attention based camera blending: 
\begin{equation}
    \mathbf{c} = softmax(\frac{QK^T}{\sqrt{d_k}} \cdot O^T)V,
\label{eq:att-rgb}
\end{equation}
where $V$ stacks all observed colors, $O$ stacks all the visibilities. Directly blending textures from source view will cause ghosting artifacts in novel view regions occluded from all source images, leading the problem ill-posed in sparse setting, as illutrated in Fig.~\ref{fig:att}. To alleviate this, we propose to regard the prediction of radiance $\mathbf{c_0}$ of Eqn.~\ref{eq:alpha-fusion} as an observation of a non-occluded virtual camera that is in identical direction with query view direction $d$. 
Thus, we could extend blending procedure in Eqn.~\ref{eq:att-rgb} to associated $N+1$ views. The associated camera blending can apply learnable blending weights in different regions across  $N+1$ views of the human performer which enables ghosting-free appearance estimation.
The final color prediction can be obtained from $\mathcal{F}_{\mathbf{c}} = T \circ \mathcal{F}_{\mathbf{c}_0}$, We call the whole process of this section as screen-space occlusion-aware appearance blending.

\subsection{Volume Rendering with Conditional Neural Radiance Field }\label{render}   

Following NeRF~\cite{mildenhall2020nerf}, we render the color of the rays passing through our implicit neural radiance filed via projective integration to obtain 2D images. Values of each rendered pixel is estimated via integrating density and colors along its corresponding camera ray $\mathbf{r}_t = \mathbf{o} + t\mathbf{\alpha}$. For a pixel $\mathbf{p}$, its corresponding camera ray $\mathbf{r}$ is calculated with known camera poses. We then uniformly sample $N$ points $\{ \mathbf{x}_{i=1}^{N} \}$ along the ray within near and far bounds $t_n$, $t_r$, which are determined according to SMPLx regressions. The expected color $C(\mathbf{r})$ is calculated as:
\begin{equation}
C(\mathbf{r}) = \int_{t_n}^{t_f} \exp \left( -\int_{t_n}^t \sigma(\mathbf{r}(s) ds \right) \sigma \  \mathbf{c} \  dt 
\end{equation}

\begin{figure*}[!t]

\centering
\includegraphics[width=.98\linewidth]{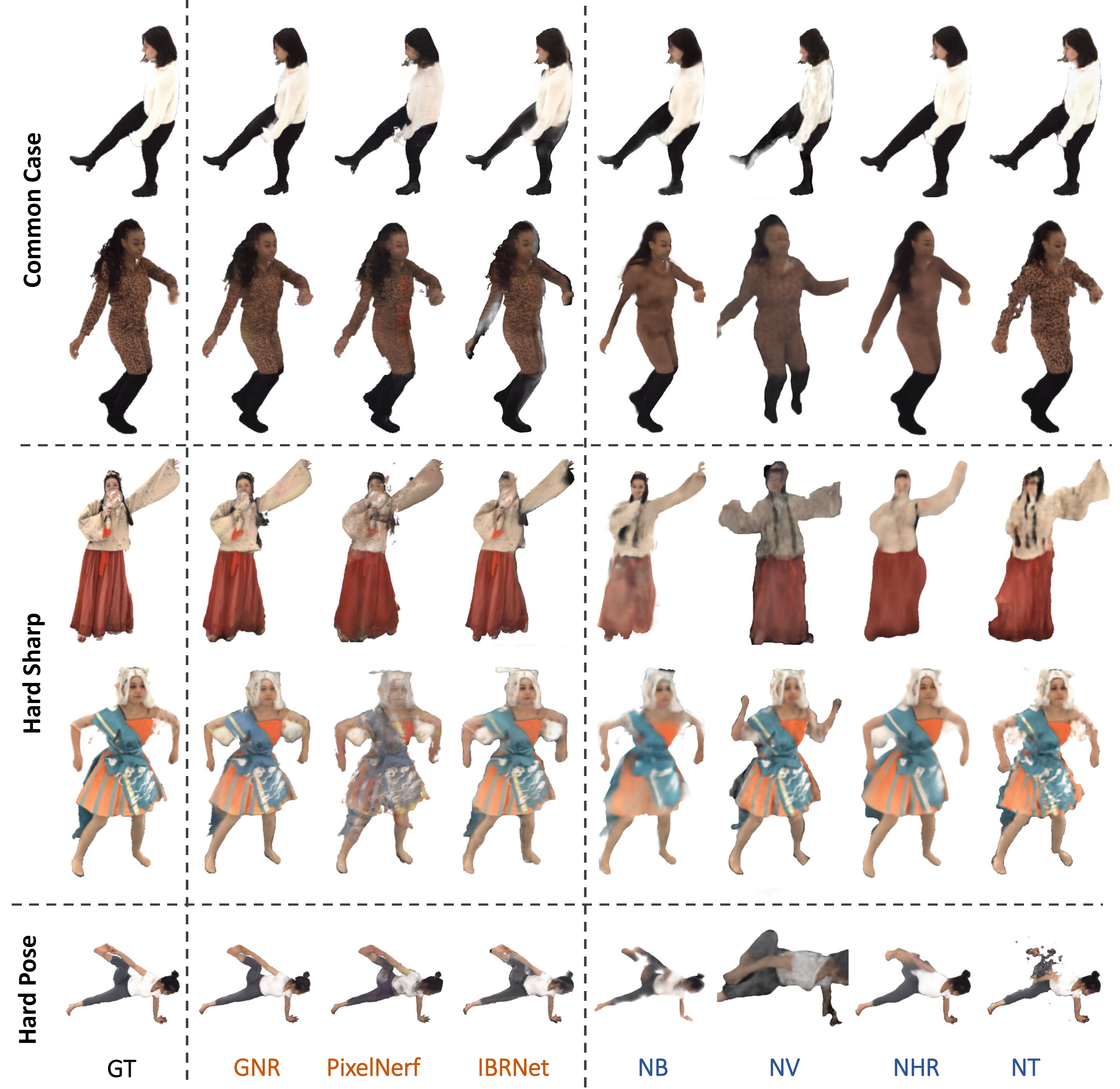}
\vspace{-1ex}
\caption{\small \textbf{Novel view synthesis on GeneBody testset under unseen ID and poses.} We compare qualitative results of novel view of unseen ID and unseen poses on generalization methods (\textcolor{orange}{Orange}) and case-specific methods (\textcolor{blue}{Blue}). Subjects in each row come from Normal Case (Kicking, Samba), Hard Shape (Cosplay, Hanfu) and Hard Pose (Yoga) subgroups respectively.
Please enlarge for details. }\label{fig:ghr-zero}
\vspace{-2ex}
\end{figure*}

\subsection{Loss Function}\label{loss}

\noindent\textbf{Photometric Loss.} We render the color of each ray using both $\mathbf{c}_0$ and blending results $\mathbf{c}$, and define photometric loss as the mean squared error (MSE) between the rendered colors and groundtruth pixel colors $\tilde{\mathbf{c}}$ for training:
\begin{equation}
    \mathcal{L}_{photo} = \sum_{\mathbf{r}\in\mathcal{R}}\left [ ||\mathbf{c}_0(\mathbf{r})-\tilde{\mathbf{c}}(\mathbf{r})||_2^2 + ||\mathbf{c}(\mathbf{r})-\tilde{\mathbf{c}}(\mathbf{r})||_2^2\right ] / |\mathcal{R}|
\vspace{-1ex}
\end{equation}
where $\mathcal{R}$ is the set of rays in the batch, and $|\mathcal{R}|$ represents the number of rays.

\noindent\textbf{Occupancy and Occlusion Loss.}
When 3D groundtruth is available, we first supervise the occupancy probability for each point in the casting rays. We model occupancy probability as $tanh(\sigma)$. Besides, we also guide the occlussion-aware network to learn reasonable 'diffused' occlusion $o$ from data, using groundtruth sreen-space occlusion $\psi(z_{gt}-z)$, where $z_{gt}$ is groundtruth rasterized depth from source view. Our occupancy and occlusion loss can be writen as:
\begin{equation}
\begin{split}
    \mathcal{L}_{geo} = & \sum_{\mathbf{r}\in\mathcal{R}}\sum_{\mathbf{x}\in\mathbf{r}} ||tanh(\sigma(\mathbf{x}))-\tilde{sgn}(\mathbf{x})||^2/|\mathcal{X}| \\ + & \sum_{\mathbf{r}\in\mathcal{R}}\sum_{\mathbf{x}\in\mathbf{r}} ||o-\psi(z_{gt}-z))||^2/|\mathcal{X}|
\end{split}
\end{equation}
where $\mathbf{x}$ is sample point from ray $\mathbf{r}$, and $|\mathcal{X}|$ is the number of all sample points. 

Overall, our method is optimized by both photo-metric and geometric loss:
 \begin{equation}
    \mathcal{L} = \mathcal{L}_{photo} + \lambda_{geo}\mathcal{L}_{geo}
\end{equation}
where $\lambda_{geo} \in \{0,1\}$, and set to $1$ when the 3D scan is available.

\section{Experiments}\label{exp}

\subsection{Datasets}\label{sec:4.1}

\begin{figure}[!t]
    \centering
    \includegraphics[width=0.95\linewidth]{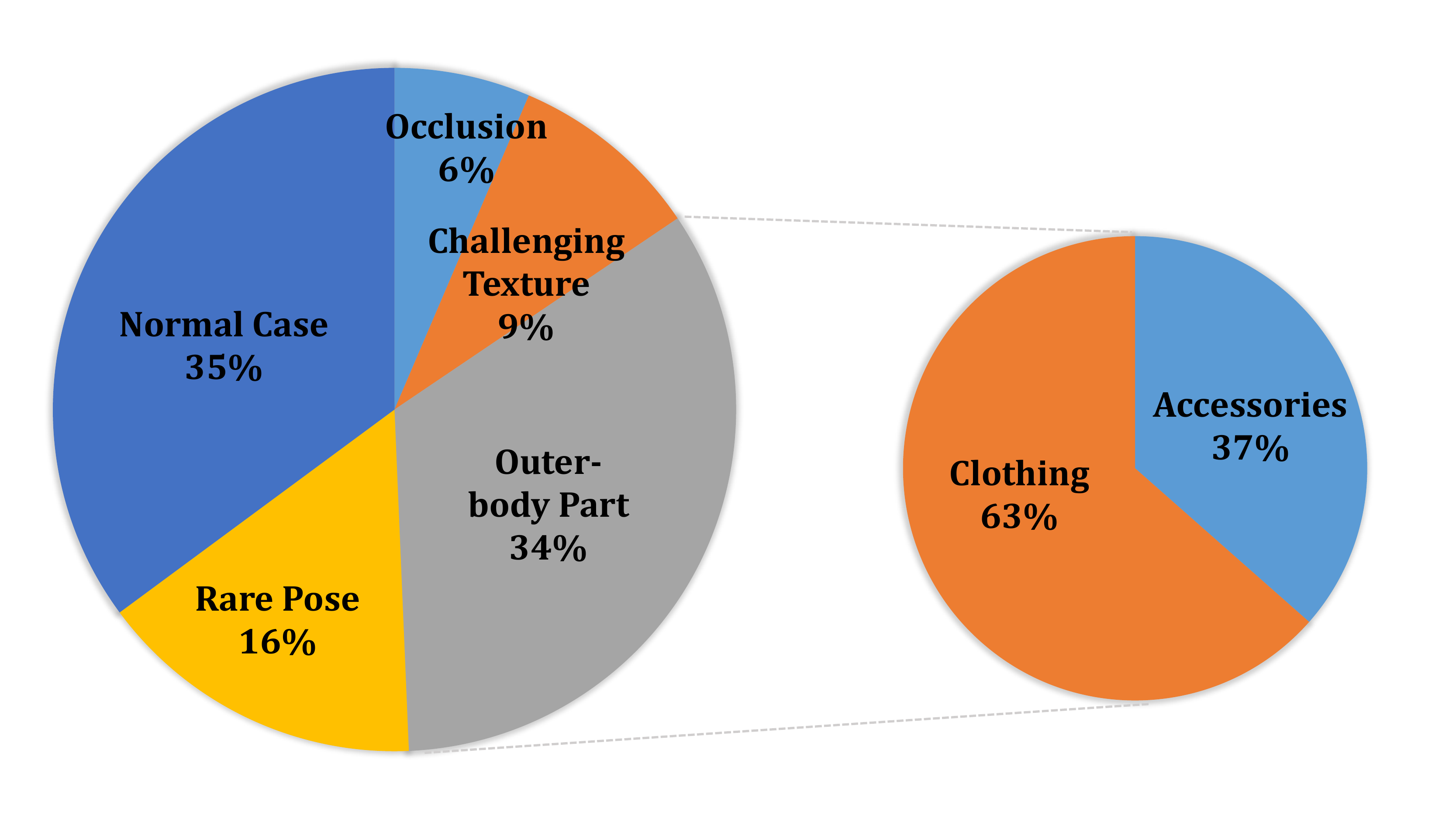}
    \caption{\small \textbf{GeneBody Statistics of Challenging Cases}.}
    \label{fig:gallery}
\end{figure}

\vspace{-0.1cm}
\noindent
\textbf{RenderPeople}~\cite{web:renderpeople}. A commercial dataset that contains high-fidelity 3D colored static scan meshes with large variety in appearance and poses. 
We acquire $260$ static human scans, and split $240$ into a train set and the rest as test set. 

\noindent
\textbf{ZJUMocap}~\cite{peng2020neural}. It contains $10$ sequences of human performances, each captured by $23$ uniformly distributed synchronized cameras. We split the first $300$ frames from three sequences as training set and rest as test set.

\noindent
\textbf{GeneBody-1.0.} GeneBody for short, is a new dataset that we collected to evaluate the generalization and robustness of human novel view synthesis. It consists of over total $2.95M$ frames of $100$ subjects performing $370$ sequences under $48$ multi-view cameras capturing, with a variety of pose actions, in different types of body shapes, clothing, accessories and hairdos, ranging the geometry and appearance  varies from everyday life to professional occasions. The SMPLx estimation and foreground segmentation are provided for each frame.
Although several groups have proposed datasets which have provided essential avenues for research into human bodies rendering, e.g., ZJUMoCap~\cite{peng2020neural} and CMU Plenoptic~\cite{Joo_2017_TPAMI}, HUMBI~\cite{yu2020humbi}. They suffer from some drawbacks as a benchmark for the generalization capability of human rendering methods. Human samples in these datasets either have limited cloth or accessory types~\cite{peng2020neural,ionescu2013human3,Xu:2018:MHP:3191713.3181973,Joo_2017_TPAMI}, small pose variance~\cite{peng2020neural,ionescu2013human3} or without daily human-object interaction~\cite{yu2020humbi}. GeneBody, on the other hand, have a broad distribution across different human races and ages as shown in Fig.~\ref{fig:distribution}. Moreover, it gathers a numerous numbers of clothing styles and poses, not only under daily life but also professional occasions such as traditional opera costume, with various outer body accessories like guitars and keyboards, 
as well as performing actions with large joint rotation like
yoga, locking dances and etc, as shown in Fig.~\ref{fig:intro-large},~\ref{fig:ghr-zero},~\ref{fig:ablate},~\ref{fig:gallery}. We select $40$ sequences as training set, and another $10$ sequences as test set in this paper. 

Unless specified, all synthesis data in RenderPeople is used by rendering around the human in a circle. For all datasets, $4$ uniformly distributed view inputs are given for all methods and metrics are evaluated in all camera views. 

\begin{table*}[h]
\vspace{-2ex}
\begin{center}
\small
\resizebox{0.95\textwidth}{!}{
\begin{tabular}{c|l|ccc|cccc}
\Xhline{0.8pt}
~&~ & \multicolumn{3}{c|}{\textbf{GeneBody-1.0}} & \multicolumn{3}{c}{\textbf{ZJUMocap}} \\
~&~ & PSNR $\uparrow$ & SSIM $\uparrow$ & LPIPS $\downarrow$ & PSNR $\uparrow$ & SSIM $\uparrow$ & LPIPS $\downarrow$ \\
\Xhline{0.8pt}
\multirow{ 4}{*}{Seen ID Unseen Pose} 
&NV~\cite{lombardi2019neural} & 19.86 & 0.774 & 0.267 & 21.74 & 0.702 & 0.285 \\
~&NHR~\cite{wu2020multi} & 20.05 & 0.800 & 0.155 & 24.79 & 0.892 & 0.185 \\
~&NT~\cite{thies2019deferred} & 21.68 & 0.881 & 0.152 & 26.38 & 0.889 & 0.146 \\
~&NB~\cite{peng2020neural} & 20.73 & 0.878 & 0.231 & 26.62 & 0.932 & 0.143 \\
\hline
\multirow{3}{*}{ Unseen ID Unseen Pose}
&pixelNerf~\cite{yu2020pixelnerf}  & 24.15 & 0.903 & 0.122 & 26.99 & 0.907 & 0.099\\
~&IBRNet~\cite{wang2021ibrnet}& 23.61 & 0.836 & 0.177
  & 28.21 & 0.921 & 0.111\\

~&\textbf{GNR} & \textbf{27.02}  & \textbf{0.931}  & \textbf{0.089}  & \textbf{28.72}  & \textbf{0.934}  & \textbf{0.081} \\
\Xhline{0.8pt}
\end{tabular}
}
\end{center}
\vspace{-2ex}
\caption{\small{\textbf{Quantitative Results on GeneBody testset.} We evaluate novel view synthesis on human performance with state-of-the-art methods on \emph{unseen ID unseen pose}. Case-specific methods are also compared on \emph{seen ID unseen pose}.}}
\vspace{-2ex}
\label{tab:g-ft-capture}
\end{table*}

\subsection{Evaluation on Real-Wold Human Performers}\label{sec:4.3}

\noindent
\textbf{Baselines and Setting.} We evaluate two categories of baseline methods:$(1)$ generalization methods, pixelNerf~\cite{yu2020pixelnerf} and IBRNet~\cite{wang2021ibrnet}; $(2)$ case-specific methods, NeuralBody (NB)~\cite{peng2020neural}, NeuralTexture (NT)~\cite{thies2019deferred}, NHR~\cite{wu2020multi} and NeuralVolumes (NV)~\cite{lombardi2019neural}. To evaluate both categories of methods on the same data, we adopt the following experimental setting.\footnote{Please check more results on other experiment settings in supplemental material.}
For generalization methods, we use the RenderPeople pretrained model (describe in Sec.~\ref{sec:4.2}) and train $5$ epochs on training set of ZJUMoCap and GeneBody separately for each dataset evaluation; and per-case models of case-specific methods are trained on first 300 frames of ZJUMoCap and first 100 frames of GeneBody in test set. 
All the rest frames are benchmarks for both categories of method for fair comparison. Such setting leads to \emph{unseen ID unseen pose} for generalization methods and \emph{seen ID unseen pose} for case-specific method.
Moreover, we use SMPLx UV map as input for NT~\cite{thies2019deferred}, and densely samples textured SMPLx for NHR's~\cite{wu2020multi} point cloud input.

\noindent \textbf{Results. } As shown in Tab.~\ref{tab:g-ft-capture} and Fig.~\ref{fig:ghr-zero}, GNR achieves the leading performance rendering novel views without any further unseen human finetuning on both human performance dataset.
NV~\cite{lombardi2019neural} takes multi-view images as the input of autoencoders and estimate mixture of affine transformation from first canonical frame, it fails to estimate the correct warpping when unseen body pose differs largely from seen ones, see $Samba$ and $Cosplay$ in Fig.~\ref{fig:ghr-zero}. NHR~\cite{wu2020multi} and NT~\cite{thies2019deferred} both uses human geometry proxies as input and render the images with a convolution network, they tend to fail when warp fields are improperly estimated which frequently happens in stretch cases, see $Hanfu$ and $Yoga$. The most related NB~\cite{peng2020neural} extracts latent code from body model vertices, it tends to learn in-precise latent code in non-rigid region with large displacement over SMPLx model,  see $Hanfu$ and $Cosplay$. Note that pixelNerf can be regard as an degradation of our method with image feature condition only, the estimated radiance usually chromaticly different from real image.

\begin{figure}[!h]
\centering
\includegraphics[width=\columnwidth]{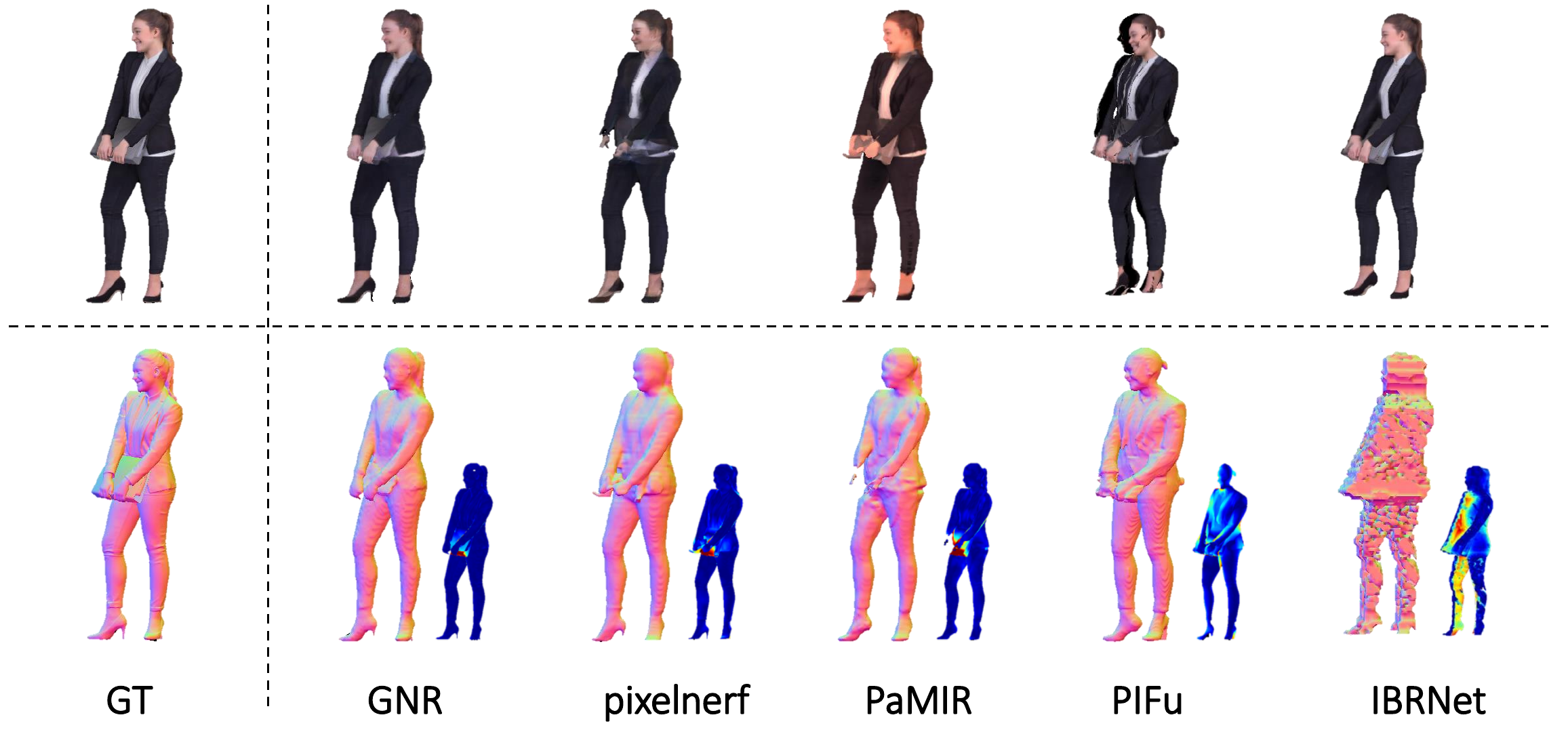}
\vspace{-4ex}
\caption{\small \textbf{Qualitative Results on RenderPeople.} Novel view images (top row), geometry (pink in bottom row) and projection error (blue in bottom row) are displayed.}\label{fig:g-ft-synth}
\vspace{-2ex}
\end{figure}

\begin{table*}[!h]

\begin{center}
\scriptsize
\resizebox{0.95\textwidth}{!}{
\begin{tabular}{l|cccc|ccc}
\Xhline{0.8pt}
~ & \multicolumn{4}{c|}{\textbf{3D Geometry Reconstruction}} & \multicolumn{3}{c}{\textbf{Novel View Synthesis}} \\
~ & Chamfer $\downarrow$ & Normal $\downarrow$ & UHD $\downarrow$ & F-score $\uparrow$ & PSNR $\uparrow$ & SSIM $\uparrow$ & LPIPS $\downarrow$ \\
\Xhline{0.8pt}
PIFu~\cite{saito2019pifu} & 6.432 & 0.606 & 22.254 & 0.445 & 22.654 & 0.865 & 0.147 \\
PIFuHD~\cite{saito2020pifuhd} & 2.765 & 0.421 & 12.621 & 0.670 & 24.386 & 0.884 & 0.096 \\
PaMIR~\cite{zheng2021pamir} & 1.565 & \textbf{0.125} & \textbf{3.585} & 0.809 & 24.870 & 0.902 & 0.080 \\
pixelNerf~\cite{yu2020pixelnerf} & 1.250 & 0.383 & 8.842 & 0.610 & 25.463 & 0.889 & 0.122 \\
IBRNet~\cite{wang2021ibrnet} & 24.864&  0.853 & 74.960  & 0.062 & 27.258  & 0.938 & 0.083 \\
\Xhline{0.8pt}
\textbf{GNR} & \textbf{0.550} & 0.198 & 4.879 & \textbf{0.807} & \textbf{29.201} & \textbf{0.964} & \textbf{0.043}\\
\Xhline{0.8pt}
\end{tabular}
}
\end{center}
\vspace{-3ex}
\caption{\small{\textbf{Quantitative Results on RenderPeople testset.} We evaluate geometry reconstruction and novel view synthesis with state-of-the-art methods.}}
\vspace{-2ex}
\label{tab:g-ft-synth}
\end{table*}

\subsection{Reconstruction and Rendering Evaluation on Synthetic Data}\label{sec:4.2}
\noindent\textbf{Baselines and Setting.} We evaluate static human generalization in two scopes, reconstruction and rendering. State-of-the-art generalizable human reconstruction methods PIFu~\cite{saito2019pifu}, PIFuHD~\cite{saito2020pifuhd} and PaMIR~\cite{zheng2021pamir}, and volume rendering methods pixelNerf~\cite{yu2020pixelnerf} and IBRNet~\cite{wang2021ibrnet} are selected as comparison baselines.
We re-train all the networks on RenderPeople training set until convergence and evaluate both quantitative results on test set in both 2D and 3D metrics. Specifically, chamfer distance, chamfer normal distance, universal Hausdorff distance (UHD), and F-score are measured between reconstructed mesh and groundtruth scan. Image quality is also estimated in PSNR, SSIM and LPIPS~\cite{zhang2018perceptual}, we obtain PIFu, PIFuHD, PaMIR's image from test views via rasterizing the output mesh.

\noindent \textbf{Results.} As shown in Tab.~\ref{tab:g-ft-synth} and Fig.~\ref{fig:g-ft-synth}, GNR achieves outstanding geometry reconstruction performance, and outperforms all methods in rendering quality. Compared with with PaMIR~\cite{zheng2021pamir} who also utilizes parametric model conditioning, GNR demonstrates its better overall geometry alignment ability (chamfer distance) for body and outer-body accessory because of geometry consistency across different views. On the other hand, despite other radiance based methods usually produce relative noisy object surface, GNR suffers much less due to contribution of body shape embedding. 
An interesting found is despite IBRNet~\cite{wang2021ibrnet} achieves high-fidelity rendering especially in close views, it provides extremely poor geometry reconstruction, even if a visual hull assumption is made.

\begin{table*}[!t]
\begin{center}
\resizebox{0.95\textwidth}{!}{
\begin{tabular}{l|cccc|ccc|ccc}
\Xhline{0.8pt}
~ & \multicolumn{7}{c|}{\textbf{Render People}} & \multicolumn{3}{c}{\textbf{GeneBody-1.0}} \\
~ & \multicolumn{4}{c|}{\textbf{3D Geometry Reconstruction}} & \multicolumn{3}{c|}{\textbf{Novel View Synthesis}} & \multicolumn{3}{c}{\textbf{Novel View Synthesis}} \\
~ & Chamfer $\downarrow$ & Normal $\downarrow$ & UHD $\downarrow$ & F-score $\uparrow$ & PSNR $\uparrow$ & SSIM $\uparrow$ & LPIPS $\downarrow$ & PSNR $\uparrow$ & SSIM $\uparrow$ & LPIPS $\downarrow$ \\
\Xhline{0.8pt}
Ours \textit{w/o} body. & 1.043 & 0.281 & 9.692 & 0.725 & 27.167 & 0.946 & 0.060 & 24.33 & 0.908 & 0.114 \\
Ours \textit{w/o} att. &  0.694 &0.201 & \textbf{4.848} & 0.800 & 27.491 & 0.945 & 0.056 & 23.61 & 0.898 & 0.131 \\
Ours \textit{w/o} occ. & 0.600 &0.205 &4.949 & 0.804 & 28.769 & 0.956 & 0.045 & 26.09 & 0.911 & 0.094 \\
\hline
\Xhline{0.8pt}
Full model & \textbf{0.550} & \textbf{0.198} & 4.879 & \textbf{0.807} & \textbf{29.201} & \textbf{0.964} & \textbf{0.043} & \textbf{27.02}  & \textbf{0.931}  & \textbf{0.089} \\
\Xhline{0.8pt}
\end{tabular}
}
\end{center}
\vspace{-3ex}
\caption{\small{\textbf{Ablation Study.} Impact of our implicit body embedding, attention-based appearance blending, occlusion-aware blending techniques are studied.}}
\vspace{-4ex}
\label{tab:ablation}
\end{table*}

\subsection{Ablation Study}\label{sec:ablate}
\begin{figure}[!t]

\centering
\includegraphics[width=\columnwidth]{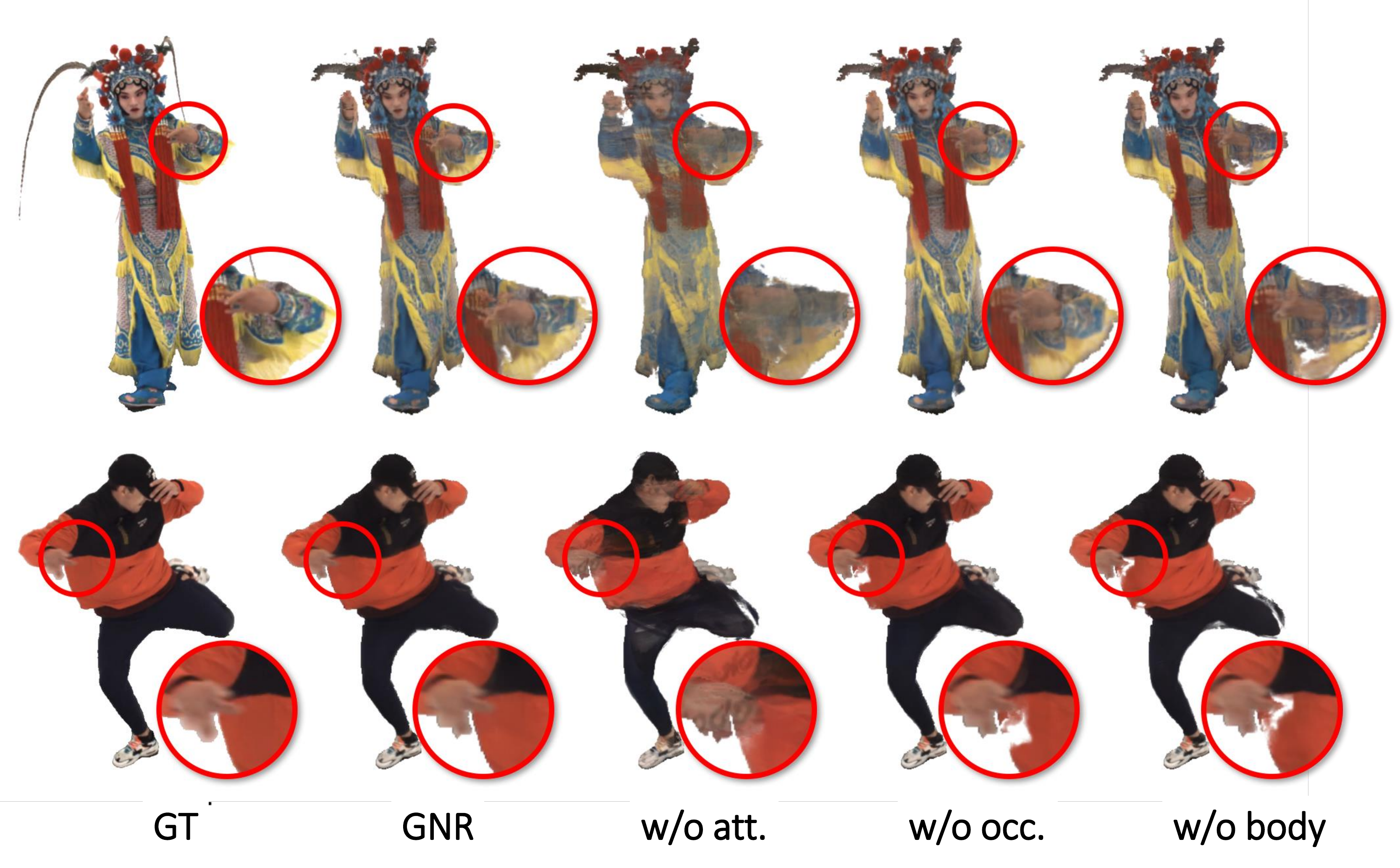}
\vspace{-4ex}
\caption{\small \textbf{Qualitative Ablation study on GeneBody Dataset.} We show the impact of proposed implicit body embedding, attention-based appearance blending, occlusion-aware blending on \textit{Peking opera, Street dance} sequence of GeneBody. Better zoom in.}\label{fig:ablate}
\vspace{-1ex}
\end{figure}

\vspace{-0.1cm}
We examine the impact of important designs on both RenderPeople and GeneBody. Settings are same as previous experiment. To examine the effectiveness of our body shape embedding, attention-based appearance blending and Screen-space Occlusion-aware appearance blending, we train three separate models that disables corresponding module, and denote them as ``\textit{w/o} body'', ``\textit{w/o} att.'', ``\textit{w/o} occ.'' respectively.
The quantitative results are shown in Tab.~\ref{tab:ablation} and Fig.~\ref{fig:ablate}.

\noindent
\textbf{Impact of Body Embedding.} The body embedding has a significant impact on the performance of geometric reconstruction. Without embedding, it failed on performers with extreme pose, and numerically, chamfer distance is reduced almost by half when the body embedding is introduced.

\noindent
\textbf{Impact of Appearance Blending.} Without attention-based appearance blending, GNR renders $\mathbf{c_0(r)}$ for each camera ray. It is shown that directly rendering out the radiance from implicit model typically fails to generate high-fidelity texture, with a degradation of $1.7-3.4$dB in PSNR. 

\noindent
\textbf{Impact of SSO Appearance Blending.} Without SSO blending the network blends from source view cameras only, ghosting texture of hands is blended to the body as shown in Fig.~\ref{fig:ablate}. SSO-AB encourages network to learn proper blending weight between observation and learned radiance from data.

\section{Conclusion}\label{discuss}
Creating free-viewpoint videos using a general model is a highly ill-posed problem due to the large shape, dressing, and pose variance of human performer. This paper contributes a dataset with  remarkable volume, which covers all aspect in general performance rendering. We also propose a novel approach GNR. Facilitated with proposed Implicit Geometric Body Embedding strategy and a Screen-Space Occlusion-Aware Blending technique, GNR successfully tailors prior knowledge of parametric model and source images, enables robust and effective estimation on human geometry and appearance. Qualitative and quantitative results demonstrate its state-of-the-art performance on multiple human datasets. 

\noindent
\textbf{Acknowledgements.} This work is supported in part by Centre for Perceptual and Interactive Intelligence Limited, in part by the General Research Fund through the Research Grants Council of Hong Kong under Grants (Nos. 14204021, 14207319, 14203118, 14208619), in part by Research Impact Fund Grant No. R5001-18, in part by CUHK Strategic Fund.

{\small
\balance
\bibliographystyle{ieee_fullname}
\bibliography{egbib}
}


\clearpage

\appendix
\noindent
\textbf{\LARGE Appendix}
\vspace{5ex}

\setcounter{equation}{0}
\setcounter{figure}{0}
\setcounter{table}{0}
\setcounter{section}{0}
\makeatletter
\renewcommand{\theequation}{S\arabic{equation}}
\renewcommand{\thefigure}{S\arabic{figure}}
\renewcommand{\thetable}{S\arabic{table}}

\vspace{-5ex}
\section{Introduction}
In this appendix, we provide detailed discussion on the design of proposed method \textit{\textbf{G}eneralizable \textbf{N}eural Performe\textbf{R}} (GNR) and additional qualitative and quantitative experiments. 

Specifically, this material is organized as follows:
$(1)$ the elaborate description of Implicit Geometric Body Embedding (Sec.~\ref{sec:embedding}); $(2)$ the detailed description of network architecture designs, training and inference strategies for GNR (Sec.~\ref{sec:imple}); $(3)$ more quantitative and qualitative results on both synthetic and real datasets(Sec.~\ref{sec:add_exp}).

\section{Implicit Body Shape Embedding}\label{sec:embedding}
Recall that GNR propose to anchor the network with a parametric dense body model, which is achieved by the implicit body shape embedding with local geometric and semantics. In this section, we provide elaborate descriptions over the implicit shape embedding stage.
Specifically, we first provide details of the body model fitting algorithm from multi-view images in Sec.~\ref{sec:fitting}. The closest point searching algorithm and gradient computation of signed distance function of our implicit body shape embedding is then discussed in Sec.~\ref{sec:closest} and in Sec.~\ref{sec:grad} respectively. 
Lastly, in Sec.~\ref{sec:grid} we explain our grid-based acceleration method and its implementation in modern GPU.

\subsection{Multi-view Body Fitting}\label{sec:fitting}
\begin{figure*}[ht!]
    \centering
    \includegraphics[width=0.98\textwidth]{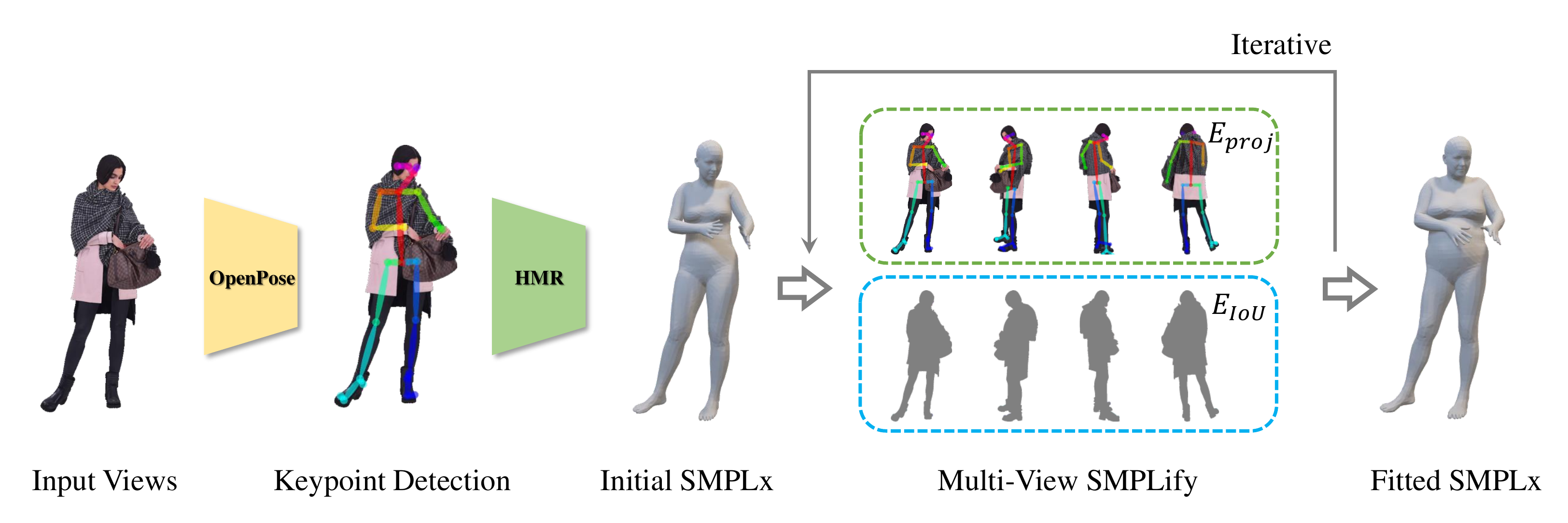}
    \vspace{-2ex}
    \caption{\small \textbf{Multi-view Body Fitting Pipeline. } 2D keypoints from multiple views are detected from OpenPose~\cite{8765346}, and the initial SMPLx model is estimated by HMR~\cite{kanazawa2018end} from one keyframe. Our multi-view fitting algorithm adopts two optimization terms, $E_{proj}$ models the skeleton reprojection error, and $E_{IoU}$ force the final SMPLx to fit tightly with shape masks. The final output is optimized in an iterative manner.}
    \label{fig:fitting}
\end{figure*}
As illustrated in Fig.~\ref{fig:fitting}, to estimate the accurate and uniform SMPLx~\cite{pavlakos2019expressive} mesh and parameters $\theta, \beta$ from multiple input views with camera matrices $R$, we first predict a coarse result as the initialization from a randomly selected input image by using ~\cite{kanazawa2018end}.
Then, we extend the SMPLify~\cite{bogo2016keep} with tailoring a multi-view joints reprojection error into the optimization:
\begin{equation}
    E_{proj} = \sum_{n=1}^{N}\sum_{k=1}^{K}\omega_{n,k}\mathbf{\rho}(\pi(sR_nJ_{n,k}(\theta,\beta)+t)-P_{n, k})
\end{equation}
where $P=\{P_k|k=1,...,K\}$ is the estimated 2D joints locations from OpenPose~\cite{8765346}, and $\omega_{n,k}$ is the corresponding confidence weight of joint $k$ in image $I_n$. $J_k$ is the 3D joint $k$ induced by $\theta, \beta$ in homogeneous coordinate. $s$ and $t$ are scale factor and global translation of the SMPLx model, finally $\mathbf{\rho}$ is the robust Geman-McClure penalty function.

Only minimizing the reprojection error may lead to misfit body shape due to the loose body shape constraint from joints location. Thus, we further define
the intersection over union (IoU) error from the projected region of SMPLx and the mask $M=\{M_n|n=1,...,N\}$ of image $I$, to encourage the inferred model fitting tighter to the given shape.
\begin{equation}
    E_{IoU} = \sum_{n}^{N} IoU(M_n, \mathbf{\Gamma}(M(\theta,\beta);\mathbf{K}, R_n))
\end{equation}
where $\mathbf{\Gamma}$ is the differentiable rendering function~\cite{kato2018renderer} given the model mesh and output the 2D mask using perspective projection, and $IoU(A,B)=|A\cap B|/|A\cup B|$.

The overall energy function can be written as 
\begin{equation}
\begin{split}
    E_{total} &= \lambda_{proj} E_{proj} + \lambda_{IoU} E_{IoU} \\
    &+ \lambda_\theta E_\theta(\theta)+ \lambda_a\theta E_a(\theta) + \lambda_\beta E_\beta(\beta) 
\end{split}
\end{equation}

$\{\lambda_{proj}, \lambda_{IoU}, \lambda_\theta, \lambda_a \lambda_\beta\}$ are trade-off parameters. $E_\theta(\theta)$, $E_a$, $E_\beta$ are regularization terms defined in~\cite{kolotouros2019spin} preventing impossible pose and shape. When the groudtruth camera poses are given, we optimize the SMPLx parameters via
\begin{equation}\label{eq:argmin_smpl}
    \theta^{\star}, \beta^{\star}, s^{\star}, t^{\star} = \argmin_{\theta, \beta, s, t}E_{total}(\theta, \beta, R)
\end{equation}
we adopt the Adam gradient decent method to solve the optimal solution until it converges to a predefined threshold.

\subsection{Closest Point Searching}\label{sec:closest}
As mentioned in the main paper, given a mesh of body model $M$ parametrized by body shape $\beta$ and pose $\theta$, we expect to find a points $\mathbf{v}$ on the mesh surface, such that it has the least distance to query point $\mathbf{x}$. This problem is identical to least distance problem to single triangular face, 
which can be formulated as a quadratic programming (QP) problem with equality and inequality constraints
\begin{equation}
\begin{split}
    \min_{c_1, c_2, c_3} & ||\sum_{i=1}^3 c_i\mathbf{v}_i - \mathbf{x}||_2  \\
    s.t \ \ \ \ & c_1, c_2, c_3 \geq 0; \\
    & c_1 + c_2 + c_3 = 1 \label{eq:closest}
\end{split}
\end{equation}
where $(\mathbf{v}_1, \mathbf{v}_2, \mathbf{v}_3)$ are vertices who construct a triangular face.
This QP problem with inequality constraint can be optimized via active set method~\cite{nocedal2006numerical}. Note that this optimization process is numerical efficient, only $4\times4$ matrix inversion is required in maximum 5 steps. Depending on optimal active set $\mathcal{A}$, there exist three types of optimal points:
\begin{itemize}
    \item When $size(\mathcal{A})=0$ or $\mathcal{A}$ is empty set, the optimal nearest point lies inside the triangular;
    \item When $size(\mathcal{A})=1$, the optimal point lies on one of the edges of the triangular;
    \item When $size(\mathcal{A})=2$, the optimal point lies on one of the vertices of the triangular;
\end{itemize}

\subsection{Gradient of Signed Distance Function}\label{sec:grad}
Our implicit body shape embedding is constructed on signed distance function (SDF) and its gradient $S'$. Recall that the SDF is defined as
\begin{align}
    S(\mathbf{x}, M) &= {\rm sign}(\mathbf{x}, M) ||\mathbf{x} - \mathbf{v}||_2,\label{eq:sdf}\\
    {\rm sign}(\mathbf{x}, M) &= \left\{\begin{matrix}
                                 1 & \mathbf{x} \  \text{inside} \  M, \\ 
                                -1 & \mathbf{x} \  \text{outside}\  M.
                                \end{matrix}\right. \label{eq:sign}
\end{align}
where $\mathbf{v}$ indicates the closest point of $\mathbf{x}$.

\begin{figure}
\centering
         \includegraphics[width=0.20\textwidth]{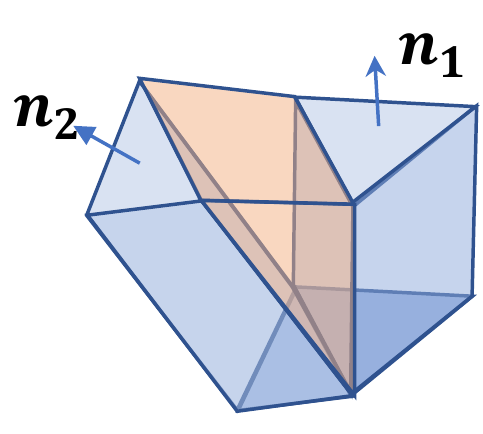}
         \hspace{0.04\textwidth}
         \includegraphics[width=0.20\textwidth]{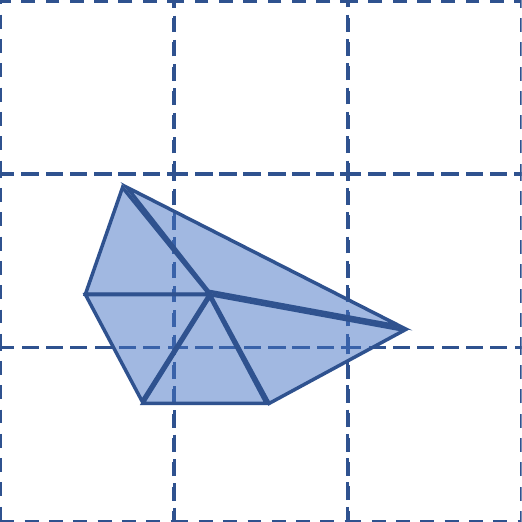}
     \caption{\small (Left) \textbf{Gradient of SDF. } Region in blue is the triangular prism, where gradient is piece-wise constant; Region in orange is the space where gradient is piece-wise smooth.
     (Right) \textbf{Voxel Grid of Mesh. } Volume is discretized into voxels, faces related to each voxel are indexed.}
  \label{fig:mesh_grid}
\end{figure}

The $S'$ can be written as
\begin{equation}
    S'(\mathbf{x}, M) = {\rm sign}(\mathbf{x}, M)\frac{(\mathbf{x}-\mathbf{v})}{||\mathbf{x}-\mathbf{v}||_2}
    \label{eq:gradient}
\end{equation}
Note that $S$ is not strictly differentialable when multiple nearest point can be found, in practice we choose the first indexed nearest point to calculate gradient.
Eqn.~\ref{eq:gradient} satisfies eikonal equation, where the gradient of SDF is a vector with norm equals to $1$. Although the defined SDF's gradient of a mesh is not a inward normal vector field due to the discretized surface, the gradient keeps smooth property , as illustrated in Fig.~\ref{fig:mesh_grid} (a). It is  piece-wise constant in the prism of a triangular face, and piece-wise smooth in the space between two adjacent prisms.

\subsection{Grid-based Hierarchical Searching}\label{sec:grid}
When computing Eqn.~\ref{eq:sdf}, naively searching from the full set of triangular faces is computational inefficient. Therefore, we utilize a grid based hierarchical searching method to narrow down the searching space. Given a input mesh $M$ of parametric body model, we first discretize the volume into voxel grids $V=\{V_i | i=1,...,m\times n\times k\}$. Integers $m,n,k$ here are resolution in three axes. Related faces which lie inside or cross each voxel are memorized as $\mathcal{P}(V)$, as shown in Fig.~\ref{fig:mesh_grid} (b). Then, we perform basic mesh operations, e.g, 
nearest point searching, ray casting, etc, in a hierarchical manner. More specifically, in the coarse stage mesh operations are performed on voxel grids, when voxel $V_i$ is visited, we test all faces in $\mathcal{P}(V_i)$ in the fine stage. 
Take closest point searching as an example, the hierarchical searching algorithm can be summarized as:

\begin{algorithm}[h]
\SetAlgoLined
 \textbf{Initialization}: Discretize the volume $V$; Compute the related faces $P(V)$\;
 Rank the non-empty voxels with Manhattan distance $d_{m}$;
$d_{min}=+\infty$; $i=0$\;
 \While{$d_{min} > \sqrt{2} d_{m}$}{
  \For{face in $\mathcal{P}(V_i)$}{
  Solve Eqn.~\ref{eq:closest}, and get distance $d=||\mathbf{x}-\mathbf{v}||_2$ \;
  \If{$d<d_{min}$}{
   $d_{min} = d$\;
   }
  }
   $i = i+1$\;
 }
 \caption{Hierarchical Searching of Closest Point}\label{alg:closest}
\end{algorithm}
To further accelerate the algorithm, we execute the process of Alg.~\ref{alg:closest} in parallel in modern GPUs. More precisely, each CUDA core optimizes the closest point with respect to one face in current voxel $V_i$. We evaluate the computational efficiency of body embedding methods in PaMIR~\cite{zheng2021pamir} and GNR in term of execution time. PaMIR requires 0.387s to extract embedding from a 3D convolution network for 64k points, while our method only takes \textbf{0.0208}s for same number of points. Note that the volume resolution is $127^3$, and execution time and memory resources required for 3D convolution increase cubically if a higher volume resolution is needed.

\section{Implementation Details}\label{sec:imple}

\subsection{Network Architecture} 
\begin{figure*}[!t]
    \centering
    \includegraphics[width=0.85\textwidth]{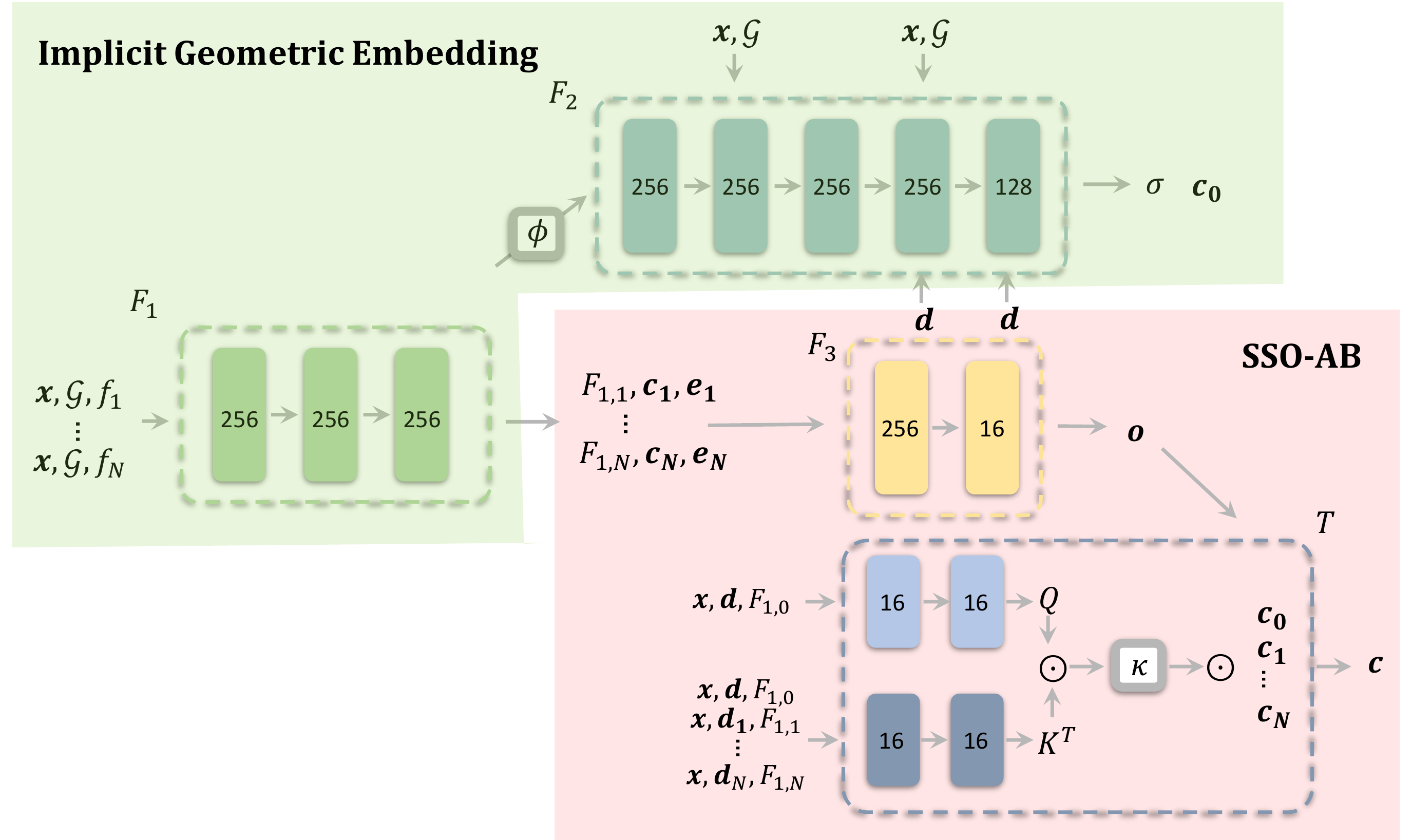}
    \caption{\small \textbf{Network Architecture Details and Data Flow. } Query point coordinate $\mathbf{x}$, its implicit body shape embedding $\mathcal{G}$ and reprojected image feature from $i$th view $f_i$ are input to $F_1$ for intermediate feature embedding; Intermediate features $F_{1,i}$ from all views are aggregated via an average pooling operator $\phi$ are then feed to occupancy prediction network $F_2$; $F_{1,i}$ are concatenated with observed color and sent to color prediction network $F_3$ after weighted pooling $\psi$.  $\mathbf{x},\mathcal{G}$, and $\mathbf{d}$ are injected to intermediate layers of $F_2$ and $F_3$ respectively. The final color is fused by learned camera blending from appearance blending network $T$.}
    \label{fig:network}
\end{figure*}

We describe detailed network architecture of our GNR, which consists of a image encoder $E$, the main implicit function who contains three MLPs $F_1, F_2, F_3$ and predicts volume density $\sigma$ and color $\mathbf{c}_0$, and appearance blending network $T$, as illustrated in Fig.~\ref{fig:network}.

\noindent\textbf{Image Encoder.} For image feature encoder $E$, stacked hourglass network is used whose parameters are initialized from the pre-trained geometry model of~\cite{saito2019pifu} and outputs a $128$-channel feature map with downsampled resolution. To resolve the overly smoothed effect due to low-resolution embedding, we then adopt a transposed convolution network to upsample feature map and produce our final image embedding in full resolution, during which the resized input image is concatenated after feature channel as an observation reference before transposed convolution at each scale. 

\noindent\textbf{Multiview Feature Fusion }
As shown in Fig.~\ref{fig:network}, the main part of implicit function consists of three sub-network $F_1, F_2, F_3$, and two pooling operations $\phi, \psi$. Specifically, for volume density prediction $\mathcal{F}_{\sigma}$, we concatenate the point coordinate, SMPLx embedding (discussed in Sec.~\ref{sec:embedding}), and image features. The concatenated inputs are fed to the MLP $F_1$ with a depth of 3 and width 256, the multi-view feature is aggregated using average pooling $\phi$. The volume density is predicted via a second 5-layer MLP, and query point coordinate with body embedding are inserted in the intermediate layers like~\cite{mildenhall2020nerf}. 
The color prediction module $\mathcal{F}_{\mathbf{c_0}}$ is a shallow MLP with a depth of 3, inputs with first 3 order spherical harmonics of view direction $\mathbf{d}$ and weighted pooled features by $\psi$.

\noindent\textbf{Appearance Blending.} The appearance blending network $T$ consists of two MLPs with width 16, the query MLP takes weighted pooled feature with point coordinate and view direction and outputs $Q$; the key MLP maps $K$ with the  intermediate features and outgoing direction. The blending weights are normalized through a softmax operator, for simplicity we denoted it as $\kappa$ in Fig.~\ref{fig:network}, and final color is predicted by camera blending of $N+1$ cameras.

\subsection{Normalized Human Coordinate and Visual Hull Bounding}

\noindent\textbf{Normalized Human Coordinate.}
To unify human body in different shapes, poses, orientations or even different scales into the same capture volume, we adopt a normalized SMPL coordinate within a range of $[-1,1)^3$, and rotated by SMPL's global orientation. The query points and view directions are transformed into normalized SMPL coordinates before feeding to the network.

\noindent\textbf{Visual Hull Bounding. }
Ray casting strategy in conventional NeRF is highly inefficient because a large amount of the samples fall in the blank space even if the searching space is bounded by a near body cylinder~\cite{cheng2018ihuman3d}. We use the visual hull~\cite{laurentini1994visual} to effectively bound the sampling space. First, binary masks are extracted from source images, and the masks are then expanded via a morphological dilation operation. Samples are then back-projected to the image planes; only points that fall within all expanded masks are treated as valid samples. Invalid points are assigned with zero volume density and zero radiance during ray integration. We also randomly sample a small number of points from the free space in each sample batch to regularize the volume density in the exteriors of the visual hull. This sampling strategy dramatically improves the training efficiency and inference speed compared to the conventional method.

\subsection{Runtime and Hardware}
We train and test GNR on 8 NVIDIA V100 GPU in parallel using PyTorch~\cite{paszke2017automatic}. In each training iteration, we randomly samples $4096$ pixels with $256$ sample points on each camera ray. The network is trained in an averaging $7.11$ iterations per second rate using multi-processing distributed data parallel.
The network converges at $20$th epoch on RenderPeople, and 3rd epoch on fine-tuning ZJUMoCap and GeneBody. For testing, GNR renders $512\times512$ images on a rate of averaging $6.25$ fps.
\section{Additional Experiments}\label{sec:add_exp}
\subsection{More Quantitative Results On Real-World Captured Datasets}

Besides the results analyzed in the main paper, we demonstrate and discuss our GNR with additional experimental settings on GeneBody-1.0 and ZJU-Mocap \textit{i.e.,} zero-shot testing and in-domain testing.

\noindent\textbf{Quantitative Results under Zero-shot Testing on GeneBody and ZJU-Mocap.} 
We add an experiment setting additionally on real-world datasets called \textit{zero-shot testing}, namely GNR and all compared methods are pretrained on RenderPeople and tested on unseen subjects in GeneBody without finetuning. This setting is designed to show the generalization ability under an even harder scenario where the domain gap between RenderPeople and GeneBody would hinder the fidelity of the view synthesis. Tab.~\ref{tab:zero-shot} shows that our method out-performs other generalizable methods on both real-world dataset.
\begin{table}[h]
\begin{center}
\small
\resizebox{\columnwidth}{!}{
\begin{tabular}{l|ccc|ccc}
\Xhline{0.8pt}
~ & \multicolumn{3}{c|}{\textbf{GeneBody}} & \multicolumn{3}{c}{\textbf{ZJU-Mocap}} \\
 ~ & PSNR $\uparrow$ & SSIM $\uparrow$ & LPIPS $\downarrow$ & PSNR $\uparrow$ & SSIM $\uparrow$ & LPIPS $\downarrow$ \\
\Xhline{0.8pt}
pixelNerf~\cite{yu2020pixelnerf} & 20.40 & 0.87 & 0.37  & 24.84  & 0.90 &  0.13\\
IBRNet~\cite{wang2021ibrnet} & 19.67 & 0.85 & 0.24  & 23.96  & 0.89 &  0.17\\
\Xhline{0.8pt}
\textbf{GNR} & \textbf{22.44}  & \textbf{0.89}  & \textbf{0.14}  & \textbf{24.96} & \textbf{0.90} & \textbf{0.11} \\
\Xhline{0.8pt}
\end{tabular}
}
\end{center}

\caption{\small{\textbf{Quantitative Results of Zero-shot Testing on GeneBody and ZJU-Mocap Datasets.} We evaluate generalizable methods on the setting of pretraining on RenderPeople dataset, without any finetuning on either real-world dataset.}}

\label{tab:zero-shot}
\end{table}

\begin{table*}[h]
\begin{center}
\small
\resizebox{\linewidth}{!}{
\begin{tabular}{l|cccccccc|cccccccc}
\Xhline{1.0pt}
~ & \multicolumn{8}{c}{\textbf{PSNR}$\uparrow$} \vline & \multicolumn{8}{c}{\textbf{SSIM}$\uparrow$} \\
~ & NV &  NT & NHR & NB & 
IBR$^\dagger$ &
IBR$^\ddagger$ &
GNR$^\dagger$ & GNR$^\ddagger$ & NV &  NT & NHR & NB &
IBR$^\dagger$ &
IBR$^\ddagger$&
GNR$^\dagger$ & GNR$^\ddagger$ \\
\Xhline{1.0pt}
 amanda & 22.64&24.13 & 25.09 &  21.81 & 19.23 &  25.13  &25.00 & 23.62 &  0.89  & 0.95  & 0.96 & 0.94 &  0.90 & 0.94 & 0.94 & 0.93\\
 barry& 23.85 & 25.57 & 19.59 & 24.43 & 25.96 & 24.31 & 25.47 & 29.28 &  0.77 & 0.89 &0.92 & 0.91&  0.88  & 0.89 & 0.90 & 0.94 \\
 fuzhizhi& 18.10 & 16.96 & 22.85 & 18.28 & 16.33 & 21.79 & 24.59 & 21.96 & 0.79 & 0.19 & 0.90 & 0.86 & 0.83 & 0.87 & 0.90 & 0.90 \\
 jinyutong& 28.21 & 18.16 & 19.68 & 19.34 &  18.45& 22.90  & 21.21& 23.90 & 0.80 & 0.85 & 0.89 & 0.88 &  0.85 & 0.90 & 0.90 & 0.90\\
joseph& 15.39 & 18.91 & 21.19 & 19.36 & 19.17 & 22.59 & 22.06 & 26.30 & 0.78& 0.87 & 0.90 & 0.89 & 0.80 & 0.90 & 0.91 & 0.94\\
maria & 15.87 & 15.51 & 19.74 & 14.16 & 16.22 & 23.14 & 22.75 & 21.51 & 0.80 & 0.82 &  0.88 & 0.81 & 0.82 & 0.89 & 0.89 & 0.90\\
mahaoran& 26.88 & 27.94 & 29.04 & 21.98 &  22.70&  23.54 & 23.61& 28.41 & 0.87 & 0.92 & 0.92 & 0.79 & 0.85 &0.79  & 0.79 & 0.93\\
natacha& 21.69 & 24.65  & 26.90 & 22.81 &  22.15  & 28.36 & 26.90& 28.71 & 0.80 & 0.88 & 0.89 & 0.90 & 0.85  & 0.90 & 0.90 & 0.91\\
soufianou& 22.72 & 26.29 & 27.87 & 24.38 &  23.25 & 28.80 & 28.15 & 27.64 & 0.83 & 0.91 & 0.92 & 0.93 & 0.87 & 0.92 & 0.93 & 0.93\\
zhuna& 13.56 & 18.66 & 19.11  & 18.82  & 15.46 & 22.56 & 22.06 & 25.40 & 0.41 & 0.88 & 0.89 & 0.87 & 0.84 & 0.88 & 0.92 & 0.93\\
\Xhline{1.0pt}
Average         &19.86 & 21.68  & 20.05 & 20.73 & 19.69 & 24.31 &\textbf{24.19} & \textbf{27.02}  & 0.77 & 0.88 & 0.80 &0.88 & 0.85 & 0.89 & \textbf{0.90}  & \textbf{0.93}\\
\Xhline{1.0pt}
\end{tabular}
}
\end{center}
\caption{\small \textbf{Result on GeneBody-1.0 in terms of PSNR and SSIM. }We compare our method with case-specific in two settings. $\dagger$ denotes $SeenID \ and \ Unseen Pose$ without any pretrain, $\ddagger$ means $UnseenID \ and \ Unseen Pose$ trained on GeneBody training test.}
\label{tab:casespecific}
\end{table*}

\noindent\textbf{Quantitative Results under In-Domain Testing.} 
We exhibit more results with two settings that under In-Domain Testing, as shown in Tab~\ref{tab:casespecific}. Specifically, the first setting (denoted as $\dagger$ in Table~\ref{tab:casespecific}) is \textit{case-specific training but testing on unseen poses}, where generalizable methods such as IBRNet~\cite{wang2021ibrnet} and ours are trained following same protocol as case-specific methods such as~\cite{peng2020neural,lombardi2019neural,thies2019deferred,wu2020multi}. This setting demonstrates the effectiveness of GNR of modeling the human geometry and appearance under a rather fair protocol comparing to case-specific methods.  The second setting (denoted as $\ddagger$ in Tab.~\ref{tab:casespecific}) is \textit{in-domain finetuned but testing on unseen ID unseen poses}, where both generalizable methods are finetuned on training set of GeneBody-1.0, and tested on the listed 10 test sequences. This setting examines the capability of generalization in the same domain. From Tab.~\ref{tab:casespecific}, GNR performs better under both settings, and is proved to benefit more than ~\cite{wang2021ibrnet} from in domain finetuning.

\begin{figure}[!t]
    \centering
    \includegraphics[width=1.12\linewidth]{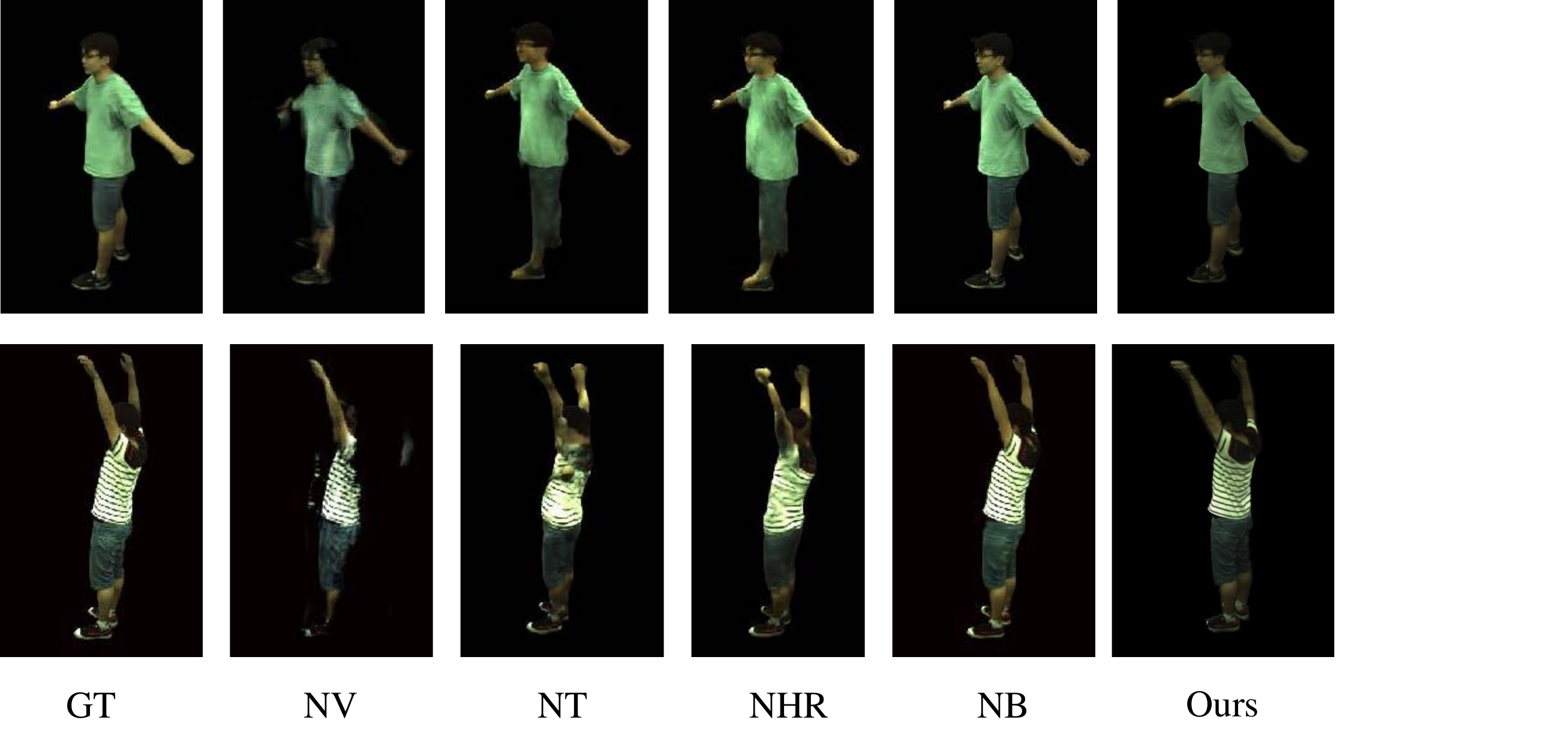}
    \caption{\small \textbf{Qualitative Results on ZJU MoCap. } We compare our methods with case-specific methods NV~\cite{lombardi2019neural}, NT~\cite{thies2019deferred}, NHR~\cite{wu2020multi} and NB\cite{peng2020neural} on this dataset. Images are from unseen camera poses of sequences in the test-set of ZJU Mocap. Our model is under the setting of training on RenderPeople and finetuning on the subset of ZJU MoCap. These two performers are unseen ID and meanwhile unseen pose to our model.}
    \vspace{-0.2cm}
    \label{fig:mocap}
\end{figure}

\subsection{More Qualitative Results On ZJU-MoCap}
We present qualitative results of our model compared with case-specific methods in the test-set of ZJU MoCap Dataset, as shown in the Fig~\ref{fig:mocap}. NV~\cite{lombardi2019neural} trends to esitmate wrong affine warping when human pose differs largely from seen poses. NT~\cite{thies2019deferred} and NHR~\cite{wu2020multi} use UV and point cloud as geometry proxy, while such proxy can not model the non-rigid deformation from human body, leading blurry texture. Our method achieve comparable visual result on total unseen pose with optimized NB~\cite{peng2020neural} model.

\subsection{More Quantitative and Qualitative Results On V-Sense}

\begin{figure}[h]
    \centering
    \includegraphics[width=\linewidth]{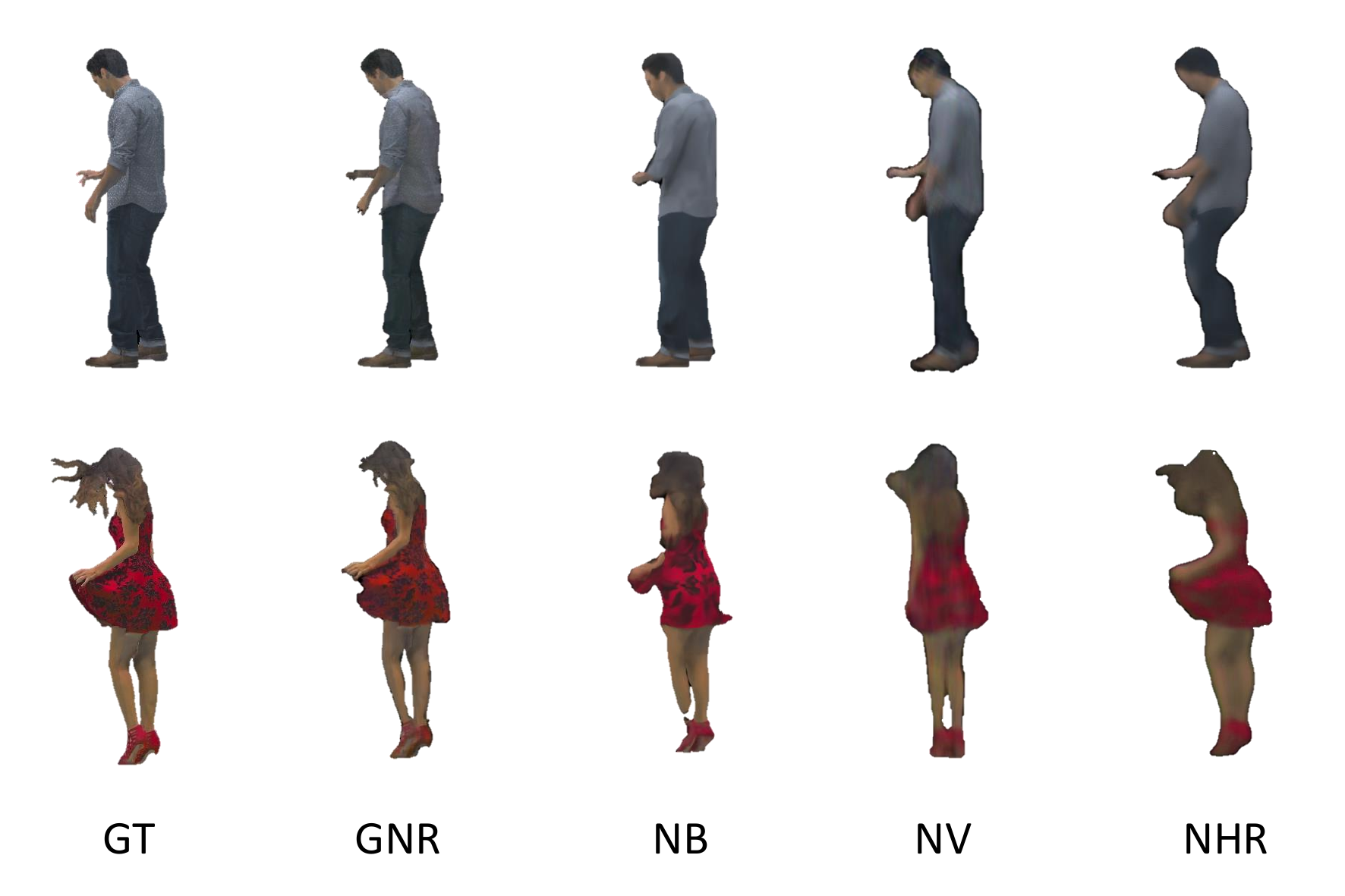}
    \caption{\small \textbf{Qualitative Results on V-Sense. } We compare our methods with case-specific methods NV~\cite{lombardi2019neural},  NHR~\cite{wu2020multi}, NB\cite{peng2020neural}.}
    \label{fig:vsense}
\end{figure}

We compare our method to case-specific methods such as~\cite{peng2020neural,lombardi2019neural,wu2020multi} on V-Sense dataset, and provide quantitative comparison in this subsection. The experiment protocol follows that of Tab. 2 in the main paper. As shown in Tab.~\ref{tab:vsense}, our method perform better than other case-specific methods. The qualitative comparison are shown in Fig.~\ref{fig:vsense}. GNR preserves better geometry fidelity comparing to methods such as ~\cite{peng2020neural,wu2020multi} and high frequency texture than ~\cite{lombardi2019neural}.

\begin{table}[h]
\begin{center}
\small
\begin{tabular}{lccccccc}
\Xhline{1.0pt}
& PSNR $\uparrow$ & SSIM $\uparrow$ & LPIPS $\downarrow$ \\
\Xhline{1.0pt}
NV~\cite{lombardi2019neural} & 24.47 &  0.88 & 0.193   \\
NHR~\cite{wu2020multi}  & 23.39 & 0.87 & 0.191 \\
NB~\cite{peng2020neural}  & 23.00 & 0.90 & 0.186 \\
\Xhline{1.0pt}
\textbf{GNR}  &\textbf{28.77}  & \textbf{0.94} & \textbf{0.068} \\

\Xhline{1.0pt}
\end{tabular}
\end{center}
\caption{\small{\textbf{Quantitative results on V-Sense.} }}
\label{tab:vsense}
\end{table}

\balance

\end{document}